\documentclass[11pt]{article}

\usepackage[final]{acl}

\usepackage{times}
\usepackage{latexsym}

\usepackage[T1]{fontenc}

\usepackage[utf8]{inputenc}

\usepackage{microtype}

\usepackage{inconsolata}

\usepackage{graphicx}
\usepackage{caption}
\usepackage{times}
\usepackage{latexsym}
\usepackage{algorithm}
\usepackage{algorithmic}
\usepackage{graphicx}
\usepackage{booktabs}
\usepackage{algorithm}
\usepackage{algorithmic}
\usepackage{amsfonts,amssymb}
\usepackage{amsmath,bm}
\usepackage{subcaption}
\usepackage{multirow}
\usepackage{multicol}
\usepackage{xcolor}
\usepackage{colortbl}
\definecolor{Gray}{gray}{0.93}
\usepackage{mathrsfs}
\usepackage{tabularx}
\usepackage{makecell}
\usepackage{amsfonts}
\usepackage{bbm}
\usepackage{arydshln}
\usepackage{bbding}
\usepackage{pifont}
\usepackage{CJKutf8}
\usepackage{tcolorbox}
\usepackage{enumitem}
\newcommand{\err}[1]{$_{\pm {#1}}$}

\title{SARA: Unlocking Multilingual Knowledge in Mixture-of-Experts via Semantically Anchored Routing Alignment}

\author{
  \textbf{Tianyu Dong\textsuperscript{1}},
  \textbf{Yangyang Liu\textsuperscript{2}},
  \textbf{Jiang Zhou\textsuperscript{1}},
  \textbf{Xinwei Wu\textsuperscript{1}},
  \textbf{Xiaohu Zhao\textsuperscript{2}},
  \textbf{Hao Wang\textsuperscript{2}},\\
  \textbf{Heng Liu\textsuperscript{2}},
  \textbf{Linlong Xu\textsuperscript{2}},
  \textbf{Longyue Wang\textsuperscript{2}},
  \textbf{Weihua Luo\textsuperscript{2}},
  \textbf{Shaolin Zhu\textsuperscript{1}\footnotemark[1]},
  \textbf{Deyi Xiong\textsuperscript{1}}
\\
\\
  \textsuperscript{1}TJUNLP Lab, School of Computer Science and Technology, Tianjin University, China\\
  \textsuperscript{2}Alibaba Group, China\\
  \texttt{\{tydong, zhushaolin, dyxiong\}@tju.edu.cn}
}

\begin{document}
\maketitle
\renewcommand{\thefootnote}{\fnsymbol{footnote}}
\footnotetext[1]{Corresponding author.}
\begin{abstract}
Sparse Mixture-of-Experts (MoE) architectures have emerged as an increasingly influential paradigm as they offer a strategic balance between parameter scalability and computational efficiency. However, low-resource languages—which suffer from a scarcity of high-quality training data— often have their tokens routed to different experts than those predominantly activated by high-resource inputs, which limits cross-lingual expert sharing. This cross-lingual routing divergence consequently hinders their efficacy in multilingual contexts. To address this issue, we propose \textbf{SARA} (Semantically Anchored Routing Alignment), a framework designed to transfer specialized capabilities from high-resource languages as anchors to low-resource languages. SARA explicitly aligns the routing distribution of multilingual inputs with high-resource semantic anchors using a symmetric Jensen-Shannon (JS) divergence constraint. Unlike traditional distillation methods that operate on output logits, SARA directly aligns the internal routing distributions of MoE layers, encouraging mechanistic consistency in expert selection across languages. We conduct experiments on 2 LLMs across 5 low-resource languages and 3 benchmarks. Experiment results demonstrate that SARA outperforms standard instruction tuning (e.g., +0.8\% on Qwen3-30B-A3B and +1.2\% on Phi-3.5-MoE-instruct on Global-MMLU benchmark). Further analyses show that SARA effectively addresses performance bottlenecks in low-resource languages, providing a scalable pathway to enhance multilingual capabilities in sparse architectures. Our code is available at https://github.com/iMoriton/sara.
\end{abstract}

\section{Introduction}
The paradigm of LLMs has shifted toward sparse MoE architectures, as evidenced by recent open-weight milestones such as Mixtral \cite{jiang2024mixtral}, DeepSeek-V3 \cite{liu2024deepseek}, DeepSeek-R1 \cite{guo2025deepseek} and Qwen3 \cite{yang2025qwen3}.
By decoupling model capacity from computational cost, MoE models achieve remarkable scalability and allow distinct subsets of parameters or experts to specialize in specific domains \cite{dai2024deepseekmoe}.
However, the efficacy of MoE primarily relies on routers correctly dispatching tokens to the most competent experts.
While these models show exceptional capability in their dominant training languages (e.g., English), extending their specialized prowess to a broader spectrum of low-resource languages remains a formidable challenge \cite{imani2023glot500, etxaniz2024multilingual, zhu2024multilingual}.

Empirical analyses of the internal mechanisms of MoE reveal a fundamental bottleneck: cross-lingual routing divergence where semantically equivalent inputs in different languages trigger disparate expert activation pathways \cite{bandarkar2025multilingual}.
Recent studies have identified the existence of super experts \cite{su2025unveiling}, defined as a sparse subset of parameters responsible for encoding sophisticated domain knowledge such as mathematical reasoning. However, these components are predominantly optimized for the data-rich patterns of high-resource languages.
When the model processes semantically equivalent inputs in low-resource languages, the routing network often struggles to generalize.
Due to surface-level lexical variations, the router fails to dispatch tokens to these high-competence experts and instead directs them toward generalist or irrelevant pathways \cite{chi2022representation}.
As a result, the model may possess the necessary parametric knowledge, but lack the routing logic required to activate it for low-resource language inputs.
Such routing misalignment not only degrades reasoning performance in low-resource languages, but also disrupts the consistency of internal representations across languages.
As observed by \citet{bandarkar2025multilingual}, this inconsistency leads to performance limitations where the hidden states of low-resource inputs diverge from their high-resource counterparts due to disparate expert composition.

Recent evaluations highlight that existing LLMs still encounter significant performance plateaus across diverse and nuanced multilingual scenarios \cite{zhang2026mitigating, chen2025benchmarking}.
Prior efforts to enhance multilingualism in LLMs have largely focused on continual pre-training or instruction tuning \cite{li2025rethinking, zhu2024landermt, li202510m}.
While recent efforts have explored expert pruning \cite{zhang2025diversifying} or load balancing optimization \cite{guo2025advancing} to improve efficiency,  they primarily optimize the routing mechanism for computational throughput rather than cross-lingual semantic consistency.
Recent works \cite{zhou2025moe, dong2025mlas, zhu2025overcoming} regulate updates via routing priors or parameter detection, revealing that unconstrained training conversely leads to catastrophic forgetting.
Consequently, the core challenge remains unaddressed: how to encourage semantically equivalent inputs to trigger similar expert activation pathways across languages?

To bridge this gap, we propose \textbf{S}emantically \textbf{A}nchored \textbf{R}outing \textbf{A}lignment (\textbf{SARA}), a novel and statistics-driven framework designed to align expert activation patterns across languages.
Our method leverages the robust routing distributions of high-resource languages as semantic anchors to rectify the routing behaviors observed in low-resource settings.
Unlike traditional distillation targeting output logits, SARA explicitly minimizes the distributional discrepancy in the routing probability space.
This encourages the routing logic to remain invariant to the input language, effectively transferring the model's capabilities to low-resource languages.
We implement this via a multi-stage pipeline: (1) constructing a semantically aligned parallel instruction corpus; (2) extracting reliable routing priors from dominant languages; (3) fine-tuning via a Jensen-Shannon (JS) divergence constraint that penalizes routing deviations.

Our contributions are threefold:

(i) We propose the SARA framework, which moves beyond traditional token-level distillation by treating the routing probability distribution of high-resource languages as semantic anchors.

(ii) By applying a symmetric JS divergence constraint on intermediate layers, SARA effectively leverages the model’s existing high-resource knowledge to make low-resource languages attain high-resource inference capabilities by rectifying their expert activation pathways.

(iii) We conduct a comprehensive evaluation across 2 LLMs and 3 challenging benchmarks (e.g., +0.8\% on Qwen3-30B-A3B and +1.2\% on Phi-3.5-MoE-instruct on Global-MMLU benchmark). Our results outperform standard instruction tuning, offering a scalable pathway for enhancing multilingual capabilities in sparse architectures.

\begin{figure*}[t]
    \centering
    \includegraphics[width=\textwidth]{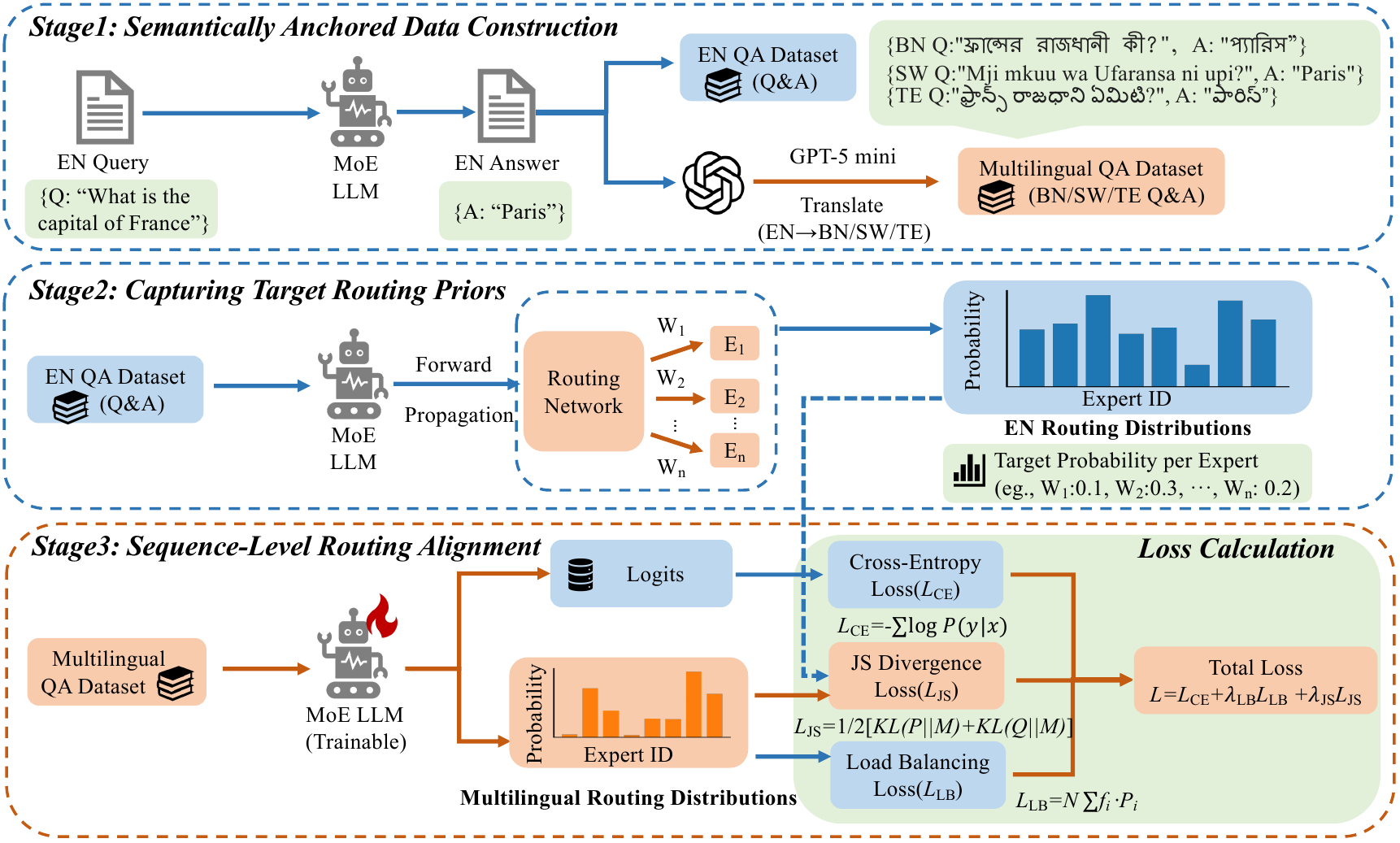}
    \caption{Illustration of the proposed SARA framework. The framework consists of three stages: (1) Generate semantically aligned parallel data via GPT-5 mini translation; (2) Perform forward propagation on high-resource inputs to extract dense routing probability distributions as target priors; (3) Fine-tune the model with a composite objective. This minimizes the Jensen-Shannon divergence between the multilingual and anchor routing distributions, along with cross-entropy and load balancing losses, to encourage cross-lingual expert consistency.}
    \label{fig:Illustration of the proposed SARA}
\end{figure*}

\section{Related Work}
\label{sec:related_work}

\paragraph{Multilingual Alignment.} Research on cross-lingual alignment in LLMs has predominantly focused on instruction fine-tuning and the strategic curation of training data.  Recent studies demonstrate that the specific composition of supervised fine-tuning (SFT) data and the integration of parallel instruction corpora \cite{penedo2025fineweb2} are pivotal for enhancing functional consistency across languages. Frameworks such as Lingualift \cite{zhang2024lingualift} introduce effective two-stage instruction tuning strategies specifically tailored to bolster performance in low-resource tasks. Aya \cite{ustun2024aya} and JetMoE \cite{shen2024jetmoe} further emphasize that distilling low-resource language knowledge from teacher models can bridge the performance gap of student models in low-resource languages. Additionally, adapting models to severely underrepresented languages highlights the critical necessity of large-scale, language-specific data curation combined with continual pre-training \cite{pan2025advancing}. While these methods demonstrate substantial improvements, they are fundamentally data-centric and mainly rely on the availability of high-quality parallel instructions and expert-curated multilingual corpora.

\paragraph{Efficiency-Oriented MoE Architectures.}Recent advances in MoE have focused on improving the quality of specialized parameter subsets through diverse initialization and hybrid designs. SCoMoE~\cite{zeng2023scomoe} optimizes MoE training efficiency by restructuring communication patterns to alleviate expensive all-to-all operations. Jamba \cite{lieber2024jamba} and ERNIE 4.5 \cite{ernie2025technicalreport} explore architectural fusion and heterogeneous scaling to optimize inference efficiency. Meanwhile, research into dynamic routing has sought to refine expert selection density based on task difficulty \cite{huang2024harder, yang2024xmoe} and mitigate computational redundancy via expert pruning \cite{lu2024not}. However, these optimizations are primarily designed for general-purpose efficiency and lack specialized mechanisms for multilingual tasks.

\paragraph{Routing Distribution Adjustments.} A bottleneck in the multilingual capability of MoE is cross-lingual routing divergence \cite{bandarkar2025multilingual}. Semantically equivalent inputs in different languages are often routed to distinct experts. This phenomenon undermines representational consistency and generalization across languages. Although existing routing-centric frameworks such as HyperMoE \cite{zhao2024hypermoe} and RoMA \cite{li2025routing} have explored adjustments to routing distribution, their objectives are largely confined to task-specific knowledge transfer or intra-domain generalization rather than cross-lingual semantic alignment. Consequently, they do not resolve the routing mismatch that occurs when moving from high-resource to low-resource linguistic contexts. Our SARA implements internal routing distillation specifically for multilingual consistency. By utilizing the model's own routing distributions on high-resource languages, SARA encourages mechanistic consistency across the routing distribution.

\section{Proposed Framework}
\label{sec:Proposed Framework}

To address the bottlenecks arising from routing misalignment in MoE, we propose the SARA framework. The key idea is to treat the expert activation patterns of high‑resource languages, as the ground‑truth signal for semantic processing. As illustrated in Figure \ref{fig:Illustration of the proposed SARA}, SARA first constructs strictly semantic‑aligned parallel data to ensure content consistency. Then we perform offline forward propagation to extract the routing distributions of high‑resource languages as priors. Finally, a distribution alignment objective is applied to pull the routing distributions of multilingual inputs toward those of their anchor texts. This encourages semantically equivalent inputs to induce similar expert activation distributions at the routing level, bridging the cross-lingual gap.

\subsection{Semantically Anchored Data Construction}
\label{sec:stage1}
The goal of SARA is to align the routing logic of low-resource languages with that of high-resource anchors. To establish a reliable semantic foundation for this anchoring process, we construct a parallel instruction dataset that captures the model’s intrinsic inference logic in high-resource contexts:

\begin{itemize}[leftmargin=*]

    \item \textbf{Correctness-based Filtering:} We first perform inference on high-resource training sets (e.g., English or Chinese). As detailed in Appendix \ref{app:prompt_templates} (referencing A.6), we require the model to enclose the final answer or option within a \texttt{\textbackslash boxed\{\}}. To objectively verify the results, we utilize the regular expression \texttt{r"$\backslash$boxed\{(.*?)\}"} to extract all formatted contents from the model's output. We designate the final element in the resulting list as the predicted answer. Only samples where the model generates a correct answer are retained. These verified samples serve as the standard for extracting internal routing patterns.

    \item \textbf{Parallel Corpus Synthesis:} For each verified sample, we translate the entire interaction (including the prompt and the logical reasoning steps) into target low-resource languages. This ensures that the low-resource samples are semantically identical to the high-resource anchor samples. By maintaining strict content consistency, the discrepancy in routing can be primarily attributed to linguistic variation, providing a clean supervision signal for alignment.
\end{itemize}

\subsection{Capturing Target Routing Priors}
\label{sec:stage2}

We extract routing distributions from high-resource data to obtain the optimal routing strategy for cross-lingual alignment.
 We perform forward propagation on the concatenated sequence of the prompt and its verified response to capture the complete inference process.
We adopt the standard sparse MoE architecture.
Given a token representation $\mathbf{h} \in \mathbb{R}^d$ (where $d$ is the hidden dimension), a gating network parameterized by $\mathbf{W}_g \in \mathbb{R}^{d \times N}$ (with $N$ denoting the number of experts) computes the routing scores $u$ and the probability distribution $\mathbf{P}$:

\begin{equation}
\mathbf{u} = \mathbf{h} \cdot \mathbf{W}_g,
\end{equation}
\begin{equation}
\mathbf{P} = \text{Softmax}(\mathbf{u}).
\end{equation}
MoE models typically use a top-$k$ selection strategy to improve efficiency. Let $\mathcal{T}$ denote the set of indices for the top-$k$ probabilities, and $\mathbf{E}_j$ represent the $j$-th expert network. The sparse output $G(\mathbf{h})$ and the layer output $\mathbf{y}$ are defined as:
\begin{equation}
G(\mathbf{h})_j = \begin{cases} P_j, & \text{if } j \in \mathcal{T} \\ 0, & \text{otherwise} \end{cases},
\end{equation}
\begin{equation}
\mathbf{y} = \sum_{j \in \mathcal{T}} G(\mathbf{h})_j \cdot \mathbf{E}_j(\mathbf{h}).
\end{equation}

To obtain stable semantic routing priors, we utilize the dense routing probability distributions produced by the gating network, rather than the discrete top-$k$ expert selections. The rationale is that for low-resource languages, the discrete expert selection during inference often diverges from high-resource anchors.

For each anchor interaction $i$, let $\mathbf{S}_\text{anchor}^{(i)} = [\mathbf{x}_i; \mathbf{y}_i]$ denote the full sequence, where $\mathbf{x}_i$ represents the input prompt and $\mathbf{y}_i$ represents the verified response. We perform forward propagation on this concatenated sequence to capture the routing logic of the complete inference process. We record the softmax-normalized routing distribution $\mathbf{R}_l$ at each layer $l$ for every token $t \in \mathbf{S}_\text{anchor}^{(i)}$:
\begin{equation}
\mathbf{R}_l(\mathbf{S}_{\text{anchor}, t}^{(i)}) = \text{Softmax}\!\left(\mathbf{h}_{\text{anchor}, t}^{(l)} \cdot \mathbf{W}_g\right) \in \mathbb{R}^{N}.
\end{equation}
To obtain a stable routing distribution, we aggregate token-level routing probabilities into a sequence-level prior $\bar{\mathbf{P}}^{(l,i)}_\text{anchor}$ using validity masks $\mathbf{m}_t$:
\begin{equation}
\bar{\mathbf{P}}^{(l,i)}_\text{anchor} = \frac{1}{\sum_t \mathbf{m}_t} \sum_t \mathbf{m}_t \cdot \mathbf{R}_l(\mathbf{S}_{\text{anchor}, t}^{(i)}).
\end{equation}
The mask $\mathbf{m}_t$ filters out [PAD] tokens to prevent non-semantic routing noise from biasing the aggregated distribution.

\subsection{Sequence-Level Routing Alignment}
\label{sec:stage3}

In this stage, we fine-tune the model on multilingual inputs to align their expert activation patterns with the extracted high-resource priors. For a multilingual interaction $i$, we similarly concatenate the translated prompt $\mathbf{x}_{\text{lang}}^{(i)}$ and response $\mathbf{y}_{\text{lang}}^{(i)}$ to form $\mathbf{S}_{\text{lang}}^{(i)} = [\mathbf{x}_{\text{lang}}^{(i)}; \mathbf{y}_{\text{lang}}^{(i)}]$. We then compute its sequence-level routing profile $\bar{\mathbf{Q}}^{(l,i)}_{\text{lang}}$:
\begin{equation}
\bar{\mathbf{Q}}^{(l,i)}_{\text{lang}} = \frac{1}{\sum_t \mathbf{m}_t} \sum_t \mathbf{m}_t \cdot \mathbf{R}_l(\mathbf{S}_{\text{lang}, t}^{(i)}).
\end{equation}

\paragraph{Routing Alignment Loss ($\mathcal{L}_\text{JS}$).}
We align the multilingual profile $\bar{\mathbf{Q}}^{(l,i)}_{\text{lang}}$ with the high-resource anchor prior $\bar{\mathbf{P}}^{(l,i)}_\text{anchor}$ by minimizing the Jensen–Shannon (JS) divergence:
\begin{equation}
\mathcal{L}_\text{JS} = \frac{1}{|{L}_\text{target}|} \sum_{l=L_\text{start}}^{L_\text{end}} \text{JS}\left( \bar{\mathbf{P}}^{(l,i)}_\text{anchor} \;\|\; \bar{\mathbf{Q}}^{(l,i)}_{\text{lang}} \right),
\end{equation}
where $|{L}_\text{target}|$ is the number of selected layers, and $L_{start}$ and $L_{end}$ denote the start and end indices of the intermediate layers (e.g., layers 7 to 34 for Qwen3). The JS divergence is defined via the Kullback-Leibler (KL) divergence as:

\begin{equation}
\text{JS}(\mathbf{P} \| \mathbf{Q}) = \frac{1}{2}\text{KL}(\mathbf{P} \| \mathbf{M}) + \frac{1}{2}\text{KL}(\mathbf{Q} \| \mathbf{M}),
\end{equation}
with $\mathbf{M} = \frac{1}{2}(\mathbf{P} + \mathbf{Q})$ being the average distribution. The choice of JS divergence over the standard KL divergence is critical due to the following 2 factors:
\begin{itemize}[leftmargin=*]
    \item \textbf{Symmetric Semantic Regularization:} Unlike KL divergence, JS divergence is symmetric, providing a balanced metric for distributional similarity. Since high-resource anchor inputs and their multilingual counterparts represent different linguistic realizations of the same semantic intent, JS divergence facilitates a more stable alignment toward a shared routing distribution.
    \item \textbf{Numerical Stability in Sparse Routing:} In MoE gating, the softmax function assigns non-zero routing probabilities to all experts, but in practice, the probabilities for unselected experts are often small. KL divergence is hypersensitive to low-probability tails, leading to disproportionately large or vanishing gradients.
\end{itemize}

\paragraph{Task Loss ($\mathcal{L}_\text{CE}$).}
To maintain the model's fundamental generative and reasoning capabilities, we apply the standard cross-entropy loss for next-token prediction to the model output in the training data:
\begin{equation}
\mathcal{L}_\text{CE} = - \sum_{t=1}^{T} \log P(\mathbf{y}_t | \mathbf{x}_{<t}).
\end{equation}

\paragraph{Load Balancing Loss ($\mathcal{L}_\text{LB}$).}
We retain the load balancing loss from Switch Transformers \cite{fedus2022switch} to prevent expert collapse and ensure efficient parameter utilization, formulated as
\begin{equation}
\mathcal{L}_\text{LB} = N \sum_{i=1}^{N} f_i \cdot \bar{\mathbf{P}}_i,
\end{equation}
where $N$ is the number of experts, $f_i$ is the fraction of tokens dispatched to expert $i$, and $\bar{\mathbf{P}}_i$ denotes the average routing probability across the batch.

The final training process uses a composite objective that balances task performance, expert utilization and cross-lingual routing consistency:
\begin{equation}
\mathcal{L}_{total} = \mathcal{L}_\text{CE} + \lambda_\text{LB}\mathcal{L}_\text{LB} + \lambda_\text{JS}\mathcal{L}_\text{JS},
\end{equation}
where $\lambda_\text{JS}$ is the hyperparameter controlling the strength of the semantically anchored routing alignment.

\section{Experiments}
\label{sec: Experiments}

We extensively evaluated the effectiveness of the proposed SARA framework. We also conducted ablation studies and in-depth analysis to verify the specific contribution of routing alignment in low-resource languages.

\subsection{Datasets}

We constructed semantically anchored instruction-tuning datasets for 5 low-resource languages(hi, ne, bn, te, sw) by applying the pipeline described in Section \ref{sec:stage1} to the high-resource splits of MMLU-ProX \cite{xuan-etal-2025-mmlu} and GSM8K \cite{cobbe2021training, yumetamath}. \footnote{We note that there is ongoing discussion in the community regarding whether hi(Hindi) should be categorized as a low-resource language \cite{holtermann2024evaluating, hangya2022improving, singh2023h, dubossarsky2024strengthening, sharma2025hi, anoop2021unsupervised}.}
We established 2 independent experimental tracks using English and Chinese as high-resource anchors, respectively, to evaluate the robustness of SARA across different semantic pivots. We utilized Qwen3-30B-A3B for initial inference and capture routing distributions from samples verified for correctness. For each anchor, the filtered interactions were translated into 5 target languages using GPT-5 mini, as detailed in Appendix \ref{app:detail_languages}.

To ensure a rigorous zero-shot evaluation, we removed any training samples overlapping with the test sets of our evaluation benchmarks. Following this process, the final corpus for each independent anchor track includes 7,000 parallel samples derived from MMLU-ProX and 7,000 samples from GSM8K per target language.

We evaluate performance on 3 diverse multilingual benchmarks: Global-MMLU \cite{singh2025global}, BELEBELE \cite{bandarkar2024belebele} and MGSM \cite{shilanguage}.

\subsection{Baselines}
We compared SARA against the following baselines:(1) \textbf{Vanilla LM}: The original instruction-tuned sparse MoE model (Qwen3-30B-A3B) without any further cross-lingual alignment. (2) \textbf{FFT}: Standard supervised full fine-tuning on the same translated dataset, but without the routing alignment objective ($\mathcal{L}_\text{JS}$). (3) \textbf{AES} \cite{guo2025advancing}: It introduces an orthogonality loss to reduce representational overlap among experts and a variance loss to encourage discriminative routing decisions. (4) \textbf{ShifCon} \cite{zhang2025shifcon}: It enhances multilingual capabilities by aligning the internal representations of non-dominant languages with the dominant language subspace.

\begin{table*}[t]
\centering
\footnotesize
\setlength{\tabcolsep}{2pt}
\renewcommand{\arraystretch}{0.8}

\begin{tabular}{ll ccccccc}
\toprule
\textbf{Benchmark} & \textbf{Anchor \& Method} & \textbf{hi} & \textbf{ne} & \textbf{bn} & \textbf{te} & \textbf{sw} & \textbf{en/zh} & \textbf{Avg.} \\
\midrule

\multirow{10}{*}{\textbf{Global-MMLU}}
& Vanilla LM & 68.15 \err{0.67} & 63.20 \err{0.26} & 64.86 \err{0.63} & 57.46 \err{0.71} & 23.66 \err{0.69} & 79.65/76.97 & 59.28 \\
\cmidrule(lr){2-9}
& \multicolumn{8}{l}{\textbf{English Anchor}} \\
& \quad FFT & 73.97 \err{0.40} & 71.91 \err{0.17} & 71.72 \err{0.27} & 65.60 \err{0.70} & 59.39 \err{0.36} & 81.80 \err{0.26} & 70.73 \\
& \quad AES & 71.53 \err{0.69} & 69.52 \err{0.45} & 70.04 \err{0.71} & 63.58 \err{0.56} & 56.09 \err{0.26} & 81.03 \err{0.21} & 68.63 \\
& \quad ShifCon & 74.32 \err{0.61} & 72.25 \err{0.78} & 72.18 \err{0.56} & 65.68 \err{0.40} & 60.15 \err{0.77} & 82.12 \err{0.41} & 71.12 \\
& \quad \textbf{SARA (Ours)} & \textbf{74.66} \err{0.15} & \textbf{72.60} \err{0.59} & \textbf{72.70} \err{0.26} & \textbf{65.76} \err{0.68} & \textbf{61.00} \err{0.43} & \textbf{82.45} \err{0.44} & \textbf{71.53} \\
\addlinespace[0.5em]
& \multicolumn{8}{l}{\textbf{Chinese Anchor}} \\
& \quad FFT & \textbf{72.69} \err{0.47} & 70.38 \err{0.11} & 70.23 \err{0.62} & 64.28 \err{0.18} & 57.88 \err{0.53} & 77.20 \err{0.62} & 68.78 \\
& \quad AES & 69.85 \err{0.72} & 68.24 \err{0.67} & 68.57 \err{0.38} & 62.13 \err{0.74} & 55.09 \err{0.43} & 76.54 \err{0.64} & 66.74 \\
& \quad ShifCon & 72.55 \err{0.24} & 70.64 \err{0.25} & 70.53 \err{0.68} & 64.48 \err{0.61} & 58.65 \err{0.35} & 77.24 \err{0.69} & 69.02 \\
& \quad \textbf{SARA (Ours)} & 72.62 \err{0.59} & \textbf{70.90} \err{0.68} & \textbf{70.84} \err{0.58} & \textbf{64.68} \err{0.27} & \textbf{59.46} \err{0.48} & \textbf{77.27} \err{0.67} & \textbf{69.30} \\

\midrule
\multirow{10}{*}{\textbf{BELEBELE}}
& Vanilla LM & 82.78 \err{0.77} & 79.67 \err{0.14} & 84.89 \err{0.78} & 73.11 \err{0.73} & 56.22 \err{0.27} & 94.67/91.78 & 78.32 \\
\cmidrule(lr){2-9}
& \multicolumn{8}{l}{\textbf{English Anchor}} \\
& \quad FFT & 82.22 \err{0.69} & \textbf{82.00} \err{0.37} & \textbf{84.11} \err{0.22} & 76.56 \err{0.72} & 77.78 \err{0.71} & 95.00 \err{0.11} & 82.95 \\
& \quad AES & 79.53 \err{0.26} & 78.85 \err{0.61} & 81.54 \err{0.77} & 74.58 \err{0.49} & 73.52 \err{0.60} & 93.84 \err{0.54} & 80.31 \\
& \quad ShifCon & 82.85 \err{0.12} & 80.75 \err{0.70} & 83.58 \err{0.49} & 77.02 \err{0.37} & 77.84 \err{0.41} & 95.12 \err{0.31} & 82.86 \\
& \quad \textbf{SARA (Ours)} & \textbf{83.44} \err{0.64} & 80.89 \err{0.22} & 83.67 \err{0.17} & \textbf{77.44} \err{0.35} & \textbf{77.89} \err{0.28} & \textbf{95.22} \err{0.49} & \textbf{83.09} \\
\addlinespace[0.5em]
& \multicolumn{8}{l}{\textbf{Chinese Anchor}} \\
& \quad FFT & 80.89 \err{0.43} & \textbf{81.44} \err{0.18} & \textbf{84.44} \err{0.23} & \textbf{76.78} \err{0.66} & 75.67 \err{0.77} & \textbf{92.00} \err{0.44} & 81.87 \\
& \quad AES & 78.56 \err{0.67} & 78.03 \err{0.46} & 80.58 \err{0.20} & 73.54 \err{0.24} & 71.09 \err{0.39} & 90.55 \err{0.21} & 78.73 \\
& \quad ShifCon & 82.05 \err{0.35} & 79.45 \err{0.61} & 83.35 \err{0.30} & 76.58 \err{0.20} & 76.55 \err{0.69} & 91.82 \err{0.67} & 81.63 \\
& \quad \textbf{SARA (Ours)} & \textbf{83.22} \err{0.55} & 79.56 \err{0.66} & 83.44 \err{0.25} & 76.67 \err{0.42} & \textbf{77.44} \err{0.27} & 91.89 \err{0.31} & \textbf{82.04} \\

\midrule
\multirow{10}{*}{\textbf{MGSM}}
& Vanilla LM & -- & -- & 84.80 \err{0.24} & 77.60 \err{0.70} & 48.40 \err{0.14} & 96.00/89.20 & 79.20 \\
\cmidrule(lr){2-9}
& \multicolumn{8}{l}{\textbf{English Anchor}} \\
& \quad FFT & -- & -- & \textbf{88.40} \err{0.79} & \textbf{84.00} \err{0.45} & 80.40 \err{0.50} & 96.00 \err{0.22} & 87.20 \\
& \quad AES & -- & -- & 85.20 \err{0.37} & 78.40 \err{0.65} & 76.00 \err{0.55} & 94.40 \err{0.76} & 83.50 \\
& \quad ShifCon & -- & -- & 87.60 \err{0.18} & 80.40 \err{0.65} & 82.40 \err{0.62} & 97.20 \err{0.38} & 86.90 \\
& \quad \textbf{SARA (Ours)} & -- & -- & 88.00 \err{0.51} & 80.80 \err{0.50} & \textbf{83.20} \err{0.77} & \textbf{97.60} \err{0.60} & \textbf{87.40} \\
\addlinespace[0.5em]
& \multicolumn{8}{l}{\textbf{Chinese Anchor}} \\
& \quad FFT & -- & -- & \textbf{86.00} \err{0.58} & \textbf{83.60} \err{0.38} & 76.80 \err{0.25} & \textbf{94.80} \err{0.26} & \textbf{85.30} \\
& \quad AES & -- & -- & 82.40 \err{0.64} & 77.60 \err{0.30} & 74.00 \err{0.69} & 84.80 \err{0.52} & 79.70 \\
& \quad ShifCon & -- & -- & 85.20 \err{0.66} & 80.40 \err{0.58} & 78.40 \err{0.73} & 86.80 \err{0.68} & 82.70 \\
& \quad \textbf{SARA (Ours)} & -- & -- & 85.60 \err{0.62} & 80.80 \err{0.79} & \textbf{79.20} \err{0.70} & 87.20 \err{0.33} & 83.20 \\
\bottomrule
\end{tabular}

\caption{Performance comparison of Qwen3-30B-A3B on Global-MMLU, BELEBELE, and MGSM. Note that the evaluation language for each benchmark is aligned with the respective semantic anchor: \textit{en Anchor} models are evaluated on English test sets, while \textit{zh Anchor} models are evaluated on Chinese test sets. Results for \textbf{hi} and \textbf{ne} are omitted for MGSM as they are not supported by the benchmark.}
\label{tab:main_results}
\end{table*}

\subsection{Settings}

We implemented SARA using the PyTorch framework and fine-tune Qwen3-30B-A3B for 2 epochs with a global batch size of 256. All models were trained on 16 × NVIDIA H100 80GB GPUs, and each training session took approximately 15 hours. The learning rate was set to 2e-5 with a cosine decay scheduler. Regarding loss coefficients, the load balancing weight $\lambda_\text{LB}$ follows the model's default configuration, while the routing alignment weight $\lambda_\text{JS}$ is fixed at 1.5. We selectively applied $\mathcal{L}_\text{JS}$ to the intermediate layers ($l \in [7, 34]$) to focus the supervision signal on the model's language-agnostic semantic core (see Appendix \ref{app:layer_analysis} for details). The English-anchored and Chinese-anchored models were trained as independent instances to verify the framework's robustness across different high-resource origins. A detailed sensitivity analysis of $\lambda_\text{JS}$ and the rationale for layer selection are provided in Appendix \ref{app:lambda_js} and \ref{app:layer_analysis}. During evaluation, we reported the average performance over 3 independent runs ($\text{Top-}p = 1, \text{temperature} = 0.1$).

\subsection{Main Results}
The performance of SARA on Global-MMLU, BELEBELE and MGSM is summarized in Table \ref{tab:main_results}. Qwen3-30B-A3B shows limited cross-lingual generalization with a pronounced performance drop on low-resource languages such as Swahili (sw) and Telugu (te). FFT substantially improves multilingual performance and serves as a strong baseline. However, FFT does not explicitly address cross-lingual routing divergence.
We perform one-tailed paired t-tests comparing SARA and FFT; the detailed statistical protocol and results are provided in Appendix~\ref{app: Statistical Significance Testing}.
We further compare SARA with AES and ShifCon under 2 anchor settings. AES relies on expert specialization and regularization, but often underperforms FFT. ShifCon applies representation-level alignment and is generally competitive with FFT. However, neither method consistently matches SARA’s performance on low-resource languages. This suggests that indirect regularization or hidden-state alignment is insufficient to correct cross-lingual expert routing.

\begin{table*}[ht!]
\centering
\footnotesize
\renewcommand{\arraystretch}{0.88}
\setlength{\tabcolsep}{10pt}
\begin{tabular}{llccccccc}

\toprule
\textbf{Benchmark} & \textbf{Method} & \textbf{hi} & \textbf{ne} & \textbf{bn} & \textbf{te} & \textbf{sw} & \textbf{en} & \textbf{Avg.} \\
\midrule
\multirow{5}{*}{\textbf{Global-MMLU}}
& -g-en-s & 73.74 & 72.03 & 72.52 & \textbf{69.67} & 58.32 & 81.47 & 71.29 \\
& -q-en-a & 69.56 & 68.09 & 67.78 & 61.81 & 55.40 & 81.78 & 67.40 \\
& -q-en-r & 70.03 & 67.69 & 68.11 & 61.84 & 55.46 & 80.64 & 67.30 \\
& -q-sw-s & 70.47 & 68.20 & 68.05 & 63.01 & 55.59 & 82.13 & 67.91 \\
& \textbf{-q-en-s (Ours)} & \textbf{74.66} & \textbf{72.60} & \textbf{72.70} & 65.76 & \textbf{61.00} & \textbf{82.45} & \textbf{71.53} \\
\midrule
\multirow{5}{*}{\textbf{BELEBELE}}
& -g-en-s & 78.22 & 77.11 & 80.00 & 73.89 & 72.33 & 93.56 & 79.19 \\
& -q-en-a & 79.22 & 77.78 & 80.89 & 72.56 & 74.56 & 94.33 & 79.89 \\
& -q-en-r & 80.00 & 76.11 & 82.00 & 74.11 & 76.00 & 92.67 & 80.15 \\
& -q-sw-s & 79.11 & 79.11 & 81.56 & 73.78 & 76.44 & 95.00 & 80.83 \\
& \textbf{-q-en-s (Ours)} & \textbf{83.44} & \textbf{80.89} & \textbf{83.67} & \textbf{77.44} & \textbf{77.89} & \textbf{95.22} & \textbf{83.09} \\
\midrule
\multirow{5}{*}{\textbf{MGSM}}
& -g-en-s & - & - & 52.80 & 47.60 & 48.80 & 62.80 & 53.00 \\
& -q-en-a & - & - & 86.80 & 80.40 & 76.80 & 95.20 & 84.80 \\
& -q-en-r & - & - & 84.00 & \textbf{82.40} & 78.80 & 94.40 & 84.90 \\
& -q-sw-s & - & - & 85.60 & 79.60 & 77.20 & 95.20 & 84.40 \\
& \textbf{-q-en-s (Ours)} & - & - & \textbf{88.00} & 80.80 & \textbf{83.20} & \textbf{97.60} & \textbf{87.40} \\
\bottomrule
\end{tabular}
\caption{Ablation study on Qwen3-30B-A3B investigating the impact of routing prior sources, anchor languages and layer selection strategies. Results for \textbf{hi} and \textbf{ne} are omitted for MGSM as they are not supported by the benchmark.}
\label{tab:ablation_qwen3}
\end{table*}

We also find that results with a Chinese anchor are weaker than those with an English anchor on all methods.
This indicates less stable routing behavior when Chinese is used as the semantic pivot.
The gap largely reflects the model’s lower baseline performance in Chinese.
By directly aligning routing probability distributions between high-resource anchors and their multilingual counterparts, SARA achieves the best average performance under the English anchor.
We further validate the effectiveness of SARA on Phi-3.5-MoE-instruct \cite{abdin2024phi3technicalreporthighly} in Appendix \ref{app:phimoe}, demonstrating that explicit expert routing alignment is an effective mechanism for transferring semantic capabilities across languages.

\begin{figure*}[ht]
    \centering
    \begin{subfigure}{0.32\textwidth}
        \centering
        \includegraphics[width=\linewidth]{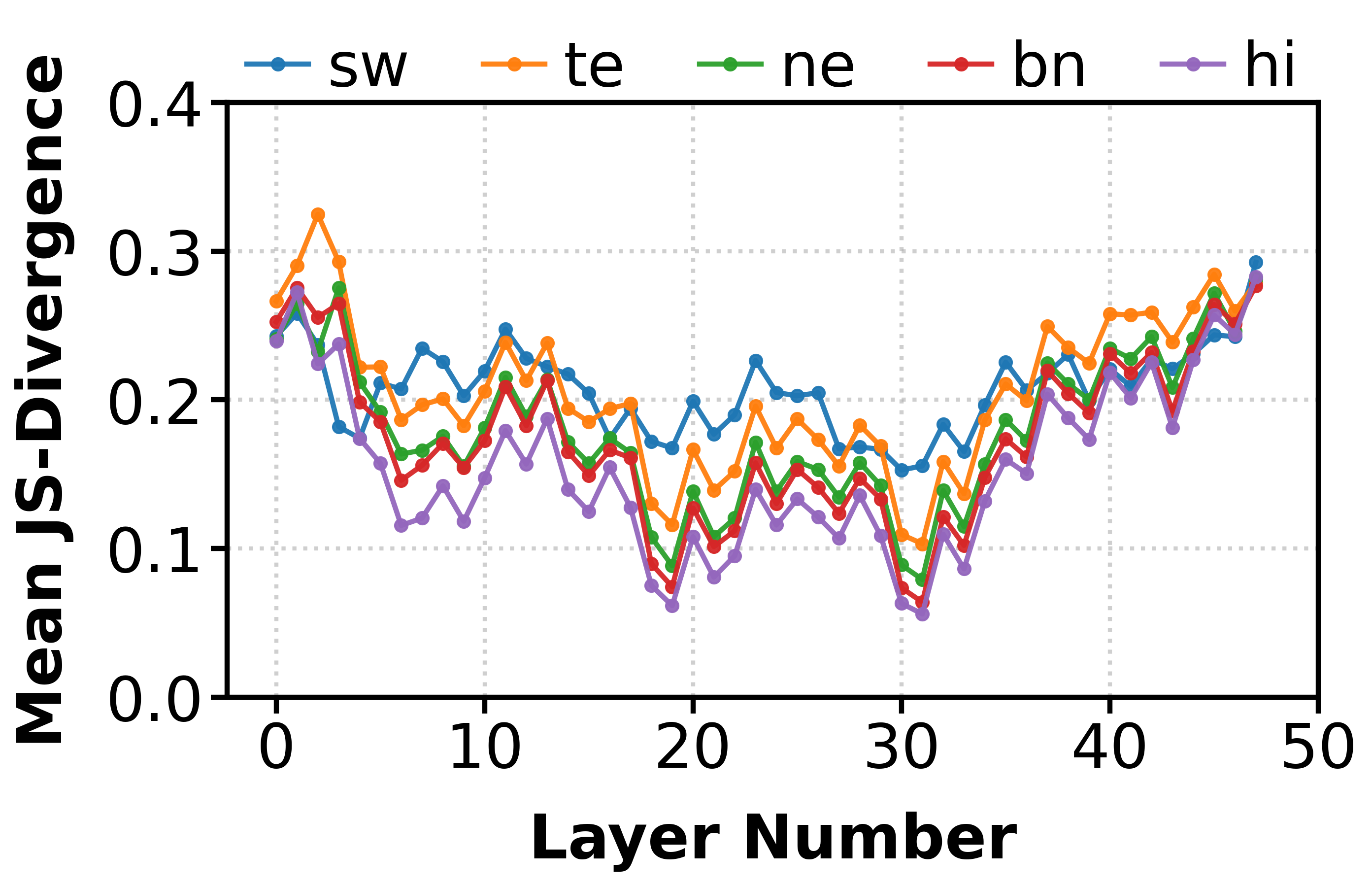}
        \caption{JS divergence without fine-tuning}
        \label{fig:routing_div}
    \end{subfigure}
    \hfill
    \begin{subfigure}{0.32\textwidth}
        \centering
        \includegraphics[width=\linewidth]{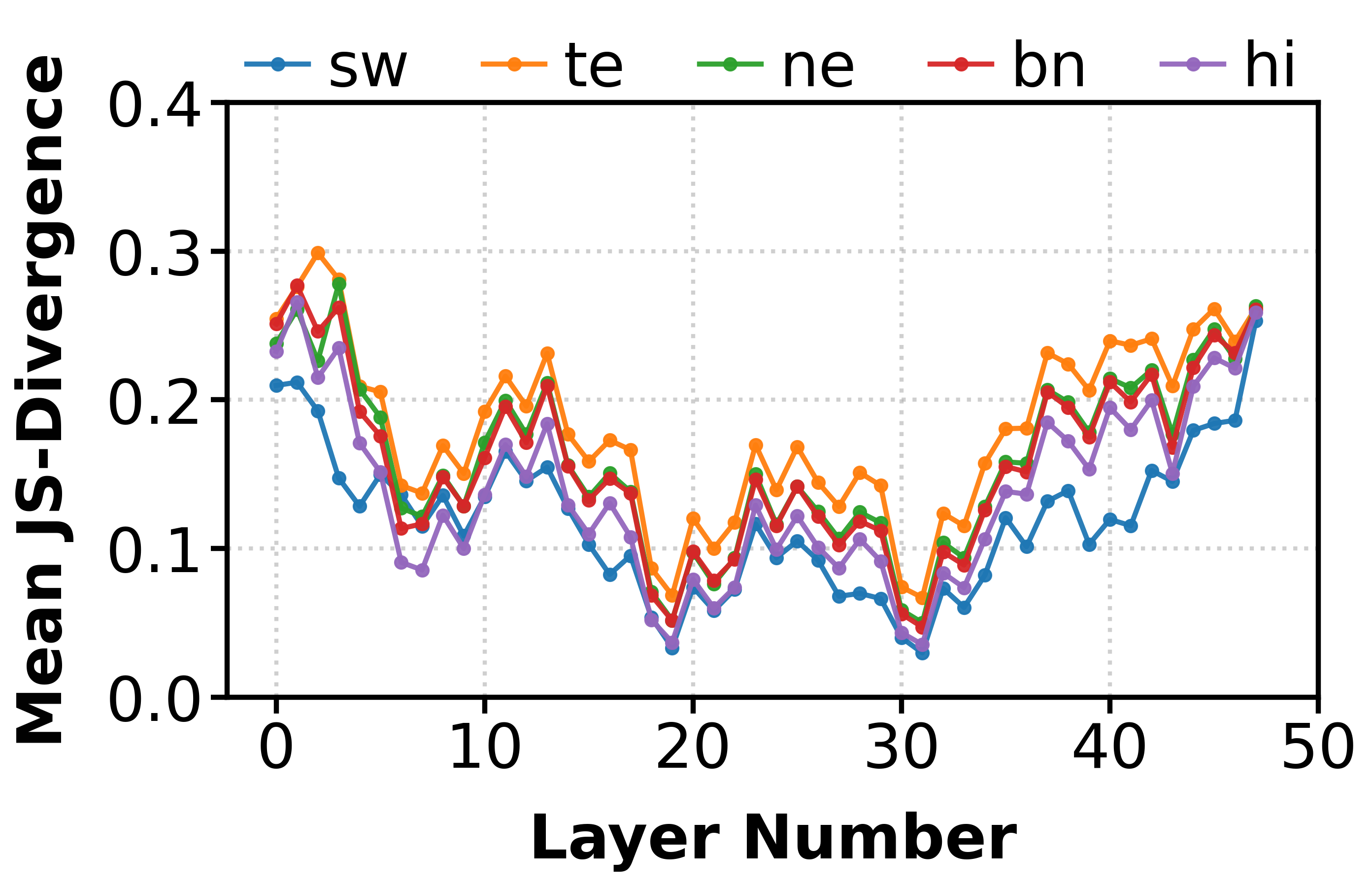}
        \caption{JS divergence after FFT}
        \label{fig:routing_div_after_sft}
    \end{subfigure}
    \hfill
    \begin{subfigure}{0.32\textwidth}
        \centering
        \includegraphics[width=\linewidth]{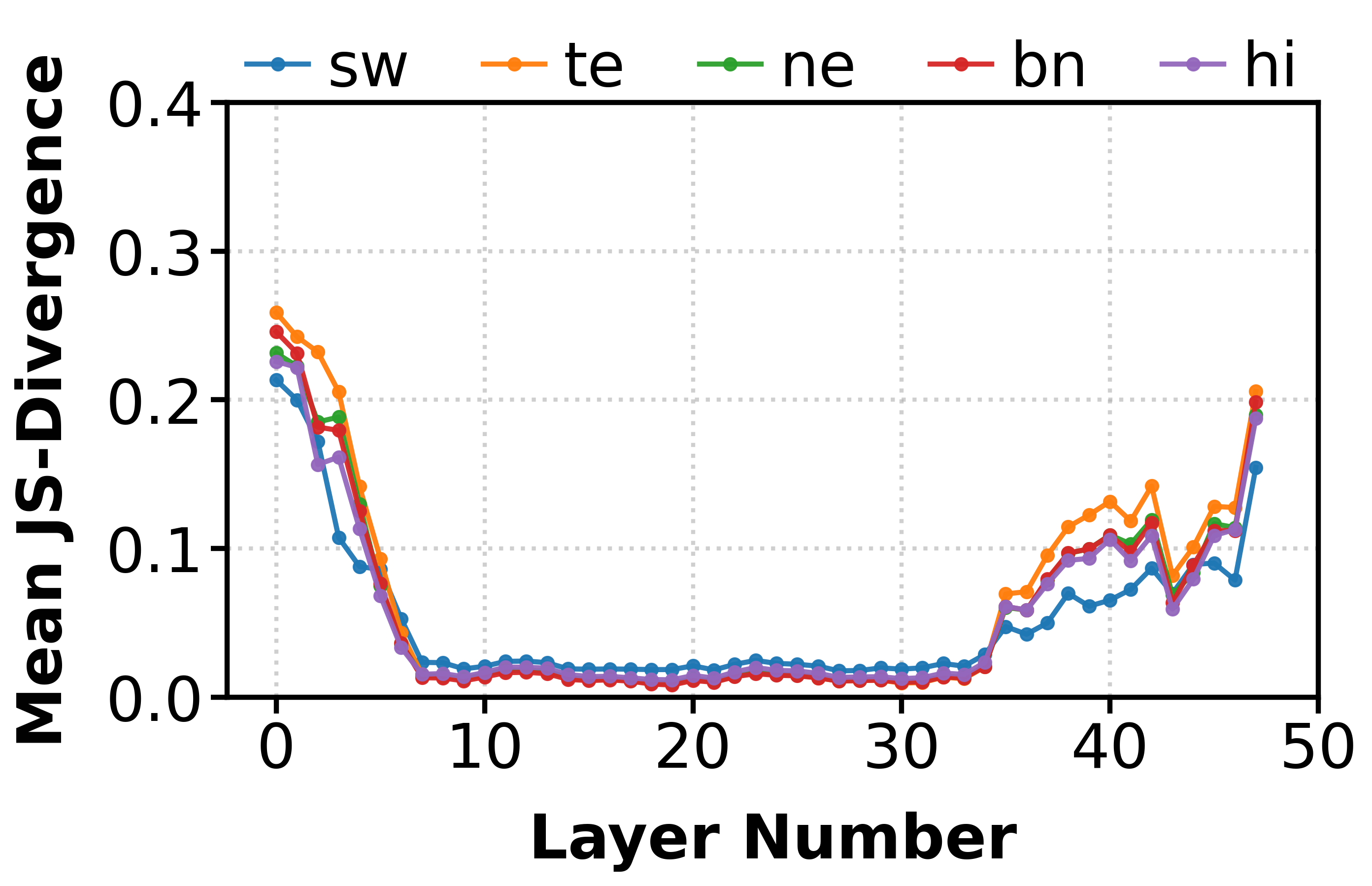}
        \caption{JS divergence after SARA}
        \label{fig:routing_div_after_sara}
    \end{subfigure}

    \caption{Comparison of layer-wise routing divergence across different fine-tuning strategies. SARA demonstrates superior alignment in intermediate layers compared to Vanilla LM and FFT.}
    \label{fig:three_figures}
\end{figure*}

\subsection{Ablation Study}
To analyze the effectiveness of SARA, we conducted a comprehensive ablation study on Qwen3-30B-A3B in Table \ref{tab:ablation_qwen3}.
Respectively, \textit{-q} and \textit{-g} denote routing priors extracted from the model's own internal inference traces and from externally generated data (via GPT-5 mini).
The suffixes \textit{-en} and \textit{-sw} indicate the choice of high-resource (English) and low-resource (Swahili) anchor languages.
For layer selection, \textit{-s} restricts routing alignment to the selected intermediate layers, while \textit{-a} applies alignment across all layers and \textit{-r} aligns a random subset of layers (see Appendix \ref{sec:ablation_variants} for details).

The comparison between -q-en-s and -g-en-s highlights the decisive advantage of self-anchoring over external distillation.
While external teachers provide high-quality text, they lack mechanistic consistency with the model's internal expert pathways.
This is most evident in the MGSM benchmark, where the -g-en-s variant suffers a substantial performance degradation.
We observe that the average MGSM score drops to 53.00\% compared to 87.40\% for our proposed method.
It confirms that rectifying routing logic via the model's own high-resource inference traces is more effective than injecting external knowledge.

Layer selection strategy also proves critical for maintaining linguistic flexibility.
On the MGSM benchmark, aligning a random subset of layers (-q-en-r) and all layers (-q-en-a) both lead to a decrease in average accuracy.
This supports our analysis that forcing alignment on shallow and deep layers disrupts the model's ability to process language-specific surface variations.
The -q-sw-s variant demonstrates that anchor quality is bounded by the model's native reasoning stability in the pivot language.
Utilizing the low-resource Swahili anchor results in an average Global-MMLU score of 67.91\% lower than the 71.53\% achieved with the English anchor.
This result shows that stable routing patterns from high-resource anchors provide more reliable priors which can help guide expert selection for low-resource inputs.

\section{Analysis}
\label{sec: analysis}
\subsection{Routing Consistency}

We calculated the JS divergence of the routing distributions for 5 languages relative to the English anchor on Global-MMLU, as shown in Figure \ref{fig:routing_div}. We further analyzed the impact of fine-tuning on this routing behavior. As shown in Figure \ref{fig:routing_div_after_sft}, FFT leads to a reduction in JS divergence compared to the base model. However, this reduction is incomplete; significant divergence remains perceptible in the intermediate layers. While FFT induces some degree of alignment as a byproduct of task supervision, it is insufficient to correct the internal expert selection logic. In contrast, SARA explicitly minimizes this discrepancy. As demonstrated in Figure \ref{fig:routing_div_after_sara}, SARA effectively suppresses the remaining divergence, flattening the curves to near-zero levels across layers 7 to 34.
This confirms that SARA effectively aligns routing distributions, encouraging semantically equivalent inputs to activate shared expert subsets (see Appendix \ref{app:layer_analysis} for details).

\begin{figure}[t]
    \centering
    \includegraphics[width=\linewidth]{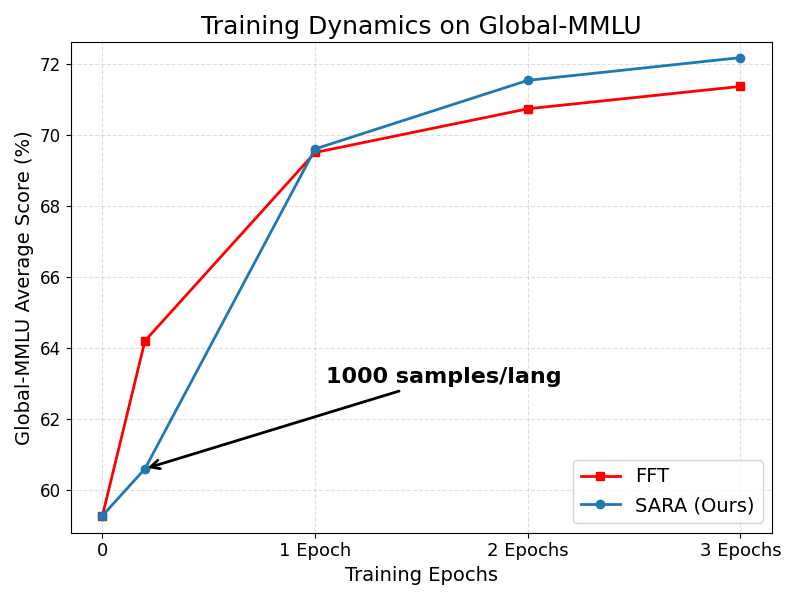}
    \caption{Training dynamics comparison between SARA and FFT on Global-MMLU.}
    \label{fig:training_dynamics}
\end{figure}

\subsection{Training Dynamics}
As shown in Figure \ref{fig:training_dynamics}, we analyzed the training dynamics on Global-MMLU to understand the specific impact of data quantity and training iterations. In the initial phase, where the model processes only 1,000 samples per language, FFT outperforms SARA. This suggests that satisfying the routing alignment constraint in addition to learning the multilingual task is more challenging than the single objective of standard fine-tuning. However, this trend reverses as the training progresses. By the end of the first epoch, SARA effectively closes the performance gap with FFT. Furthermore, as training progresses into the second epoch, the performance improvement from FFT becomes insignificant. In contrast, SARA leverages these additional training iterations to consolidate mechanistic consistency, ultimately unlocking a higher performance ceiling through internal routing. By the third epoch, both methods enter a saturation regime with only marginal gains, yet SARA consistently converges to a higher plateau than FFT, indicating a superior asymptotic performance enabled by routing consistency.

\begin{figure}[ht]
    \centering
    \includegraphics[width=1\linewidth]{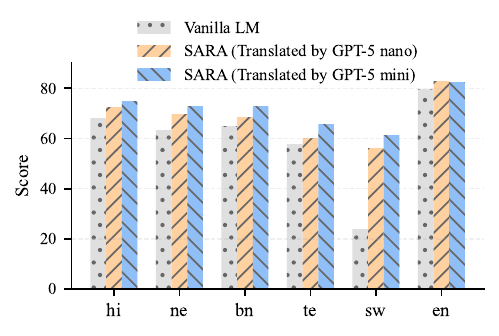}
    \caption{Comparison of SARA on Global-MMLU benchmark using translations generated by GPT-5 mini and GPT-5 nano.}
    \label{fig:translation_quality}
\end{figure}

\subsection{Effect of Translation Quality}

We analyzed the effect of translation quality on SARA by varying the models used to generate the multilingual training data. Specifically, we compared two SARA variants trained on parallel corpora translated by GPT-5 mini and GPT-5 nano, respectively.

To formally quantify the translation quality, we employed CometKiwi~\cite{rei2022cometkiwi}, a reference-free machine translation evaluation metric. As shown in Table~\ref{tab:comet_scores}, the GPT-5 mini model consistently achieves higher average CometKiwi scores across all evaluated target languages (hi, bn, te, sw, ne) compared to the GPT-5 nano model.

\begin{table}[htbp]
\centering
\resizebox{\columnwidth}{!}{
\begin{tabular}{lccccc}
\toprule
\textbf{Model} & \textbf{hi} & \textbf{bn} & \textbf{te} & \textbf{sw} & \textbf{ne} \\
\midrule
GPT-5 mini & \textbf{0.7238} & \textbf{0.7528} & \textbf{0.7356} & \textbf{0.6774} & \textbf{0.7005} \\
GPT-5 nano & 0.7179          & 0.7455          & 0.7275          & 0.6615          & 0.6982          \\
\bottomrule
\end{tabular}
}
\caption{CometKiwi Scores (en $\rightarrow$ xx) for Training Data}
\label{tab:comet_scores}
\end{table}

This disparity in translation quality directly impacts downstream performance. As illustrated in Figure~\ref{fig:translation_quality}, while both SARA variants substantially improve performance over the vanilla MoE model across all Global-MMLU languages, the variant trained with GPT-5 mini translations achieves superior accuracy. This correlation suggests that higher-quality training translations provide cleaner and more precise semantic supervision for routing alignment, thereby enhancing the model's cross-lingual transfer capabilities.

\section{Conclusion}
\label{sec: Conclusion}

In this paper, we have presented SARA, a novel framework designed to address performance bottlenecks in MoE models arising from cross-lingual routing divergence.
By treating the routing distributions of high-resource languages as semantic anchors, SARA encourages consistent expert activation patterns within intermediate layers via a Jensen-Shannon divergence constraint.
Extensive experiments conducted across multiple benchmarks demonstrate that our method effectively unlocks latent domain knowledge for low-resource languages, outperforming standard full fine-tuning.
SARA achieves these performance gains while preserving proficiency in the anchor language, thereby effectively transferring inference capabilities across linguistic boundaries.

\section*{Acknowledgments}
The present research was supported by the National Key Research and Development Program of China (Grant No. 2023YFE0116400) and National Natural Science Foundation of China Youth Fund (Grant No. 62306210).
We would like to thank the anonymous reviewers for their insightful comments.

\section*{Limitations}
First, although SARA demonstrates strong performance, the diversity of its training datasets could be improved. The model currently adapts to the specific textual styles of its training data, which may hinder generalization to distinct stylistic characteristics. Moreover, our method relies on the semantic equivalence between anchors and target languages. Translation artifacts from the synthetic generation process  could introduce noise into the routing supervision, potentially bounding the alignment performance.
Then, by explicitly aligning the routing distribution of multilingual inputs with high-resource semantic anchors, there is a risk of suppressing language-specific cultural nuances. While this effectively transfers reasoning capabilities, it may introduce a high-resource language-centric bias, limiting performance on tasks requiring deep local cultural understanding.
Finally, our layer selection strategy focuses on intermediate layers based on the observed U-shaped routing divergence in the tested models. While effective for Qwen3 and Phi-3.5, this heuristic may require adaptation for MoE architectures with fundamentally different gating mechanisms or depth configurations.

\bibliography{custom}

@article{fedus2022switch,
  title={Switch transformers: Scaling to trillion parameter models with simple and efficient sparsity},
  author={Fedus, William and Zoph, Barret and Shazeer, Noam},
  journal={Journal of Machine Learning Research},
  volume={23},
  number={120},
  pages={1--39},
  year={2022}
}

@article{liu2024deepseek,
  title={Deepseek-v3 technical report},
  author={Liu, Aixin and Feng, Bei and Xue, Bing and Wang, Bingxuan and Wu, Bochao and Lu, Chengda and Zhao, Chenggang and Deng, Chengqi and Zhang, Chenyu and Ruan, Chong and others},
  journal={arXiv preprint arXiv:2412.19437},
  year={2024}
}

@article{yang2025qwen3,
  title={Qwen3 technical report},
  author={Yang, An and Li, Anfeng and Yang, Baosong and Zhang, Beichen and Hui, Binyuan and Zheng, Bo and Yu, Bowen and Gao, Chang and Huang, Chengen and Lv, Chenxu and others},
  journal={arXiv preprint arXiv:2505.09388},
  year={2025}
}

@article{bandarkar2025multilingual,
  title={Multilingual Routing in Mixture-of-Experts},
  author={Bandarkar, Lucas and Yang, Chenyuan and Fayyaz, Mohsen and Hu, Junlin and Peng, Nanyun},
  journal={arXiv preprint arXiv:2510.04694},
  year={2025}
}

@article{guo2025advancing,
  title={Advancing Expert Specialization for Better MoE},
  author={Guo, Hongcan and Lu, Haolang and Nan, Guoshun and Chu, Bolun and Zhuang, Jialin and Yang, Yuan and Che, Wenhao and Leng, Sicong and Cui, Qimei and Jiang, Xudong},
  journal={arXiv preprint arXiv:2505.22323},
  year={2025}
}

@inproceedings{zhou2025moe,
  title={Moe-lpr: Multilingual extension of large language models through mixture-of-experts with language priors routing},
  author={Zhou, Hao and Wang, Zhijun and Huang, Shujian and Huang, Xin and Han, Xue and Feng, Junlan and Deng, Chao and Luo, Weihua and Chen, Jiajun},
  booktitle={Proceedings of the AAAI Conference on Artificial Intelligence},
  volume={39},
  number={24},
  pages={26092--26100},
  year={2025}
}

@article{jiang2024mixtral,
  title={Mixtral of experts},
  author={Jiang, Albert Q and Sablayrolles, Alexandre and Roux, Antoine and Mensch, Arthur and Savary, Blanche and Bamford, Chris and Chaplot, Devendra Singh and Casas, Diego de las and Hanna, Emma Bou and Bressand, Florian and others},
  journal={arXiv preprint arXiv:2401.04088},
  year={2024}
}

@inproceedings{dai2024deepseekmoe,
  title={DeepSeekMoE: Towards Ultimate Expert Specialization in Mixture-of-Experts Language Models},
  author={Dai, Damai and Deng, Chengqi and Zhao, Chenggang and Xu, Rx and Gao, Huazuo and Chen, Deli and Li, Jiashi and Zeng, Wangding and Yu, Xingkai and Wu, Y and others},
  booktitle={Proceedings of the 62nd Annual Meeting of the Association for Computational Linguistics (Volume 1: Long Papers)},
  pages={1280--1297},
  year={2024}
}

@article{su2025unveiling,
  title={Unveiling super experts in mixture-of-experts large language models},
  author={Su, Zunhai and Li, Qingyuan and Zhang, Hao and Ye, Weihao and Xue, Qibo and Qian, YuLei and Xie, Yuchen and Wong, Ngai and Yuan, Kehong},
  journal={arXiv preprint arXiv:2507.23279},
  year={2025}
}

@article{chi2022representation,
  title={On the representation collapse of sparse mixture of experts},
  author={Chi, Zewen and Dong, Li and Huang, Shaohan and Dai, Damai and Ma, Shuming and Patra, Barun and Singhal, Saksham and Bajaj, Payal and Song, Xia and Mao, Xian-Ling and others},
  journal={Advances in Neural Information Processing Systems},
  volume={35},
  pages={34600--34613},
  year={2022}
}

@article{li2025rethinking,
  title={Rethinking Multilingual Continual Pretraining: Data Mixing for Adapting LLMs Across Languages and Resources},
  author={Li, Zihao and Ji, Shaoxiong and Luo, Hengyu and Tiedemann, J{\"o}rg},
  journal={arXiv preprint arXiv:2504.04152},
  year={2025}
}

@inproceedings{dong2025mlas,
  title={MLAS-LoRA: Language-Aware parameters detection and LoRA-based knowledge transfer for multilingual machine translation},
  author={Dong, Tianyu and Li, Bo and Liu, Jinsong and Zhu, Shaolin and Xiong, Deyi},
  booktitle={Proceedings of the 63rd Annual Meeting of the Association for Computational Linguistics (Volume 1: Long Papers)},
  pages={15645--15660},
  year={2025}
}

@inproceedings{zhang2025diversifying,
  title={Diversifying the expert knowledge for task-agnostic pruning in sparse mixture-of-experts},
  author={Zhang, Zeliang and Liu, Xiaodong and Cheng, Hao and Xu, Chenliang and Gao, Jianfeng},
  booktitle={Findings of the Association for Computational Linguistics: ACL 2025},
  pages={86--102},
  year={2025}
}

@article{lieber2024jamba,
  title={Jamba: A hybrid transformer-mamba language model},
  author={Lieber, Opher and Lenz, Barak and Bata, Hofit and Cohen, Gal and Osin, Jhonathan and Dalmedigos, Itay and Safahi, Erez and Meirom, Shaked and Belinkov, Yonatan and Shalev-Shwartz, Shai and others},
  journal={arXiv preprint arXiv:2403.19887},
  year={2024}
}

@article{shen2024jetmoe,
  title={Jetmoe: Reaching llama2 performance with 0.1 m dollars},
  author={Shen, Yikang and Guo, Zhen and Cai, Tianle and Qin, Zengyi},
  journal={arXiv preprint arXiv:2404.07413},
  year={2024}
}

@inproceedings{ustun2024aya,
  title={Aya model: An instruction finetuned open-access multilingual language model},
  author={{\"U}st{\"u}n, Ahmet and Aryabumi, Viraat and Yong, Zheng and Ko, Wei-Yin and D’souza, Daniel and Onilude, Gbemileke and Bhandari, Neel and Singh, Shivalika and Ooi, Hui-Lee and Kayid, Amr and others},
  booktitle={Proceedings of the 62nd Annual Meeting of the Association for Computational Linguistics (Volume 1: Long Papers)},
  pages={15894--15939},
  year={2024}
}

@inproceedings{huang2024harder,
  title={Harder Task Needs More Experts: Dynamic Routing in MoE Models},
  author={Huang, Quzhe and An, Zhenwei and Zhuang, Nan and Tao, Mingxu and Zhang, Chen and Jin, Yang and Xu, Kun and Chen, Liwei and Huang, Songfang and Feng, Yansong},
  booktitle={Proceedings of the 62nd Annual Meeting of the Association for Computational Linguistics (Volume 1: Long Papers)},
  pages={12883--12895},
  year={2024}
}

@inproceedings{yang2024xmoe,
  title={XMoE: Sparse Models with Fine-grained and Adaptive Expert Selection},
  author={Yang, Yuanhang and Qi, Shiyi and Gu, Wenchao and Wang, Chaozheng and Gao, Cuiyun and Xu, Zenglin},
  booktitle={Findings of the Association for Computational Linguistics ACL 2024},
  pages={11664--11674},
  year={2024}
}

@inproceedings{lu2024not,
  title={Not All Experts are Equal: Efficient Expert Pruning and Skipping for Mixture-of-Experts Large Language Models},
  author={Lu, Xudong and Liu, Qi and Xu, Yuhui and Zhou, Aojun and Huang, Siyuan and Zhang, Bo and Yan, Junchi and Li, Hongsheng},
  booktitle={Proceedings of the 62nd Annual Meeting of the Association for Computational Linguistics (Volume 1: Long Papers)},
  pages={6159--6172},
  year={2024}
}

@inproceedings{zhao2024hypermoe,
  title={HyperMoE: Towards Better Mixture of Experts via Transferring Among Experts},
  author={Zhao, Hao and Qiu, Zihan and Wu, Huijia and Wang, Zili and He, Zhaofeng and Fu, Jie},
  booktitle={Proceedings of the 62nd Annual Meeting of the Association for Computational Linguistics (Volume 1: Long Papers)},
  pages={10605--10618},
  year={2024}
}

@article{li2025routing,
  title={Routing Manifold Alignment Improves Generalization of Mixture-of-Experts LLMs},
  author={Li, Zhongyang and Li, Ziyue and Zhou, Tianyi},
  journal={arXiv preprint arXiv:2511.07419},
  year={2025}
}

@misc{ernie2025technicalreport,
      title={ERNIE 4.5 Technical Report},
      author={Baidu-ERNIE-Team},
      year={2025},
      eprint={},
      archivePrefix={arXiv},
      primaryClass={cs.CL},
      url={https://ernie.baidu.com/blog/publication/ERNIE_Technical_Report.pdf}
}

@article{penedo2025fineweb2,
  title={FineWeb2: One Pipeline to Scale Them All--Adapting Pre-Training Data Processing to Every Language},
  author={Penedo, Guilherme and Kydl{\'\i}{\v{c}}ek, Hynek and Sabol{\v{c}}ec, Vinko and Messmer, Bettina and Foroutan, Negar and Kargaran, Amir Hossein and Raffel, Colin and Jaggi, Martin and Von Werra, Leandro and Wolf, Thomas},
  journal={arXiv preprint arXiv:2506.20920},
  year={2025}
}

@inproceedings{xuan-etal-2025-mmlu,
    title = "{MMLU}-{P}ro{X}: A Multilingual Benchmark for Advanced Large Language Model Evaluation",
    author = "Xuan, Weihao  and
      Yang, Rui  and
      Qi, Heli  and
      Zeng, Qingcheng  and
      Xiao, Yunze  and
      Feng, Aosong  and
      Liu, Dairui  and
      Xing, Yun  and
      Wang, Junjue  and
      Gao, Fan  and
      Lu, Jinghui  and
      Jiang, Yuang  and
      Li, Huitao  and
      Li, Xin  and
      Yu, Kunyu  and
      Dong, Ruihai  and
      Gu, Shangding  and
      Li, Yuekang  and
      Xie, Xiaofei  and
      Juefei-Xu, Felix  and
      Khomh, Foutse  and
      Yoshie, Osamu  and
      Chen, Qingyu  and
      Teodoro, Douglas  and
      Liu, Nan  and
      Goebel, Randy  and
      Ma, Lei  and
      Marrese-Taylor, Edison  and
      Lu, Shijian  and
      Iwasawa, Yusuke  and
      Matsuo, Yutaka  and
      Li, Irene",
    editor = "Christodoulopoulos, Christos  and
      Chakraborty, Tanmoy  and
      Rose, Carolyn  and
      Peng, Violet",
    booktitle = "Proceedings of the 2025 Conference on Empirical Methods in Natural Language Processing",
    month = nov,
    year = "2025",
    address = "Suzhou, China",
    publisher = "Association for Computational Linguistics",
    url = "https://aclanthology.org/2025.emnlp-main.79/",
    doi = "10.18653/v1/2025.emnlp-main.79",
    pages = "1513--1532",
    ISBN = "979-8-89176-332-6"
}

@article{cobbe2021training,
  title={Training verifiers to solve math word problems},
  author={Cobbe, Karl and Kosaraju, Vineet and Bavarian, Mohammad and Chen, Mark and Jun, Heewoo and Kaiser, Lukasz and Plappert, Matthias and Tworek, Jerry and Hilton, Jacob and Nakano, Reiichiro and others},
  journal={arXiv preprint arXiv:2110.14168},
  year={2021}
}

@inproceedings{yumetamath,
  title={MetaMath: Bootstrap Your Own Mathematical Questions for Large Language Models},
  author={Yu, Longhui and Jiang, Weisen and Shi, Han and YU, Jincheng and Liu, Zhengying and Zhang, Yu and Kwok, James and Li, Zhenguo and Weller, Adrian and Liu, Weiyang},
  booktitle={The Twelfth International Conference on Learning Representations}
}

@inproceedings{singh2025global,
  title={Global mmlu: Understanding and addressing cultural and linguistic biases in multilingual evaluation},
  author={Singh, Shivalika and Romanou, Angelika and Fourrier, Cl{\'e}mentine and Adelani, David Ifeoluwa and Ngui, Jian Gang and Vila-Suero, Daniel and Limkonchotiwat, Peerat and Marchisio, Kelly and Leong, Wei Qi and Susanto, Yosephine and others},
  booktitle={Proceedings of the 63rd Annual Meeting of the Association for Computational Linguistics (Volume 1: Long Papers)},
  pages={18761--18799},
  year={2025}
}

@inproceedings{bandarkar2024belebele,
  title={The belebele benchmark: a parallel reading comprehension dataset in 122 language variants},
  author={Bandarkar, Lucas and Liang, Davis and Muller, Benjamin and Artetxe, Mikel and Shukla, Satya Narayan and Husa, Donald and Goyal, Naman and Krishnan, Abhinandan and Zettlemoyer, Luke and Khabsa, Madian},
  booktitle={Proceedings of the 62nd Annual Meeting of the Association for Computational Linguistics (Volume 1: Long Papers)},
  pages={749--775},
  year={2024}
}

@inproceedings{shilanguage,
  title={Language models are multilingual chain-of-thought reasoners},
  author={Shi, Freda and Suzgun, Mirac and Freitag, Markus and Wang, Xuezhi and Srivats, Suraj and Vosoughi, Soroush and Chung, Hyung Won and Tay, Yi and Ruder, Sebastian and Zhou, Denny and others},
  booktitle={The Eleventh International Conference on Learning Representations}
}

@misc{abdin2024phi3technicalreporthighly,
      title={Phi-3 Technical Report: A Highly Capable Language Model Locally on Your Phone}, 
      author={Marah Abdin and Jyoti Aneja and Hany Awadalla and Ahmed Awadallah and Ammar Ahmad Awan and Nguyen Bach and Amit Bahree and Arash Bakhtiari and Jianmin Bao and Harkirat Behl and Alon Benhaim and Misha Bilenko and Johan Bjorck and Sébastien Bubeck and Martin Cai and Qin Cai and Vishrav Chaudhary and Dong Chen and Dongdong Chen and Weizhu Chen and Yen-Chun Chen and Yi-Ling Chen and Hao Cheng and Parul Chopra and Xiyang Dai and Matthew Dixon and Ronen Eldan and Victor Fragoso and Jianfeng Gao and Mei Gao and Min Gao and Amit Garg and Allie Del Giorno and Abhishek Goswami and Suriya Gunasekar and Emman Haider and Junheng Hao and Russell J. Hewett and Wenxiang Hu and Jamie Huynh and Dan Iter and Sam Ade Jacobs and Mojan Javaheripi and Xin Jin and Nikos Karampatziakis and Piero Kauffmann and Mahoud Khademi and Dongwoo Kim and Young Jin Kim and Lev Kurilenko and James R. Lee and Yin Tat Lee and Yuanzhi Li and Yunsheng Li and Chen Liang and Lars Liden and Xihui Lin and Zeqi Lin and Ce Liu and Liyuan Liu and Mengchen Liu and Weishung Liu and Xiaodong Liu and Chong Luo and Piyush Madan and Ali Mahmoudzadeh and David Majercak and Matt Mazzola and Caio César Teodoro Mendes and Arindam Mitra and Hardik Modi and Anh Nguyen and Brandon Norick and Barun Patra and Daniel Perez-Becker and Thomas Portet and Reid Pryzant and Heyang Qin and Marko Radmilac and Liliang Ren and Gustavo de Rosa and Corby Rosset and Sambudha Roy and Olatunji Ruwase and Olli Saarikivi and Amin Saied and Adil Salim and Michael Santacroce and Shital Shah and Ning Shang and Hiteshi Sharma and Yelong Shen and Swadheen Shukla and Xia Song and Masahiro Tanaka and Andrea Tupini and Praneetha Vaddamanu and Chunyu Wang and Guanhua Wang and Lijuan Wang and Shuohang Wang and Xin Wang and Yu Wang and Rachel Ward and Wen Wen and Philipp Witte and Haiping Wu and Xiaoxia Wu and Michael Wyatt and Bin Xiao and Can Xu and Jiahang Xu and Weijian Xu and Jilong Xue and Sonali Yadav and Fan Yang and Jianwei Yang and Yifan Yang and Ziyi Yang and Donghan Yu and Lu Yuan and Chenruidong Zhang and Cyril Zhang and Jianwen Zhang and Li Lyna Zhang and Yi Zhang and Yue Zhang and Yunan Zhang and Xiren Zhou},
      year={2024},
      eprint={2404.14219},
      archivePrefix={arXiv},
      primaryClass={cs.CL},
      url={https://arxiv.org/abs/2404.14219}, 
}

@article{guo2025deepseek,
  title={Deepseek-r1: Incentivizing reasoning capability in llms via reinforcement learning},
  author={Guo, Daya and Yang, Dejian and Zhang, Haowei and Song, Junxiao and Zhang, Ruoyu and Xu, Runxin and Zhu, Qihao and Ma, Shirong and Wang, Peiyi and Bi, Xiao and others},
  journal={arXiv preprint arXiv:2501.12948},
  year={2025}
}

@inproceedings{imani2023glot500,
  title={Glot500: Scaling Multilingual Corpora and Language Models to 500 Languages},
  author={Imani, Ayyoob and Lin, Peiqin and Kargaran, Amir Hossein and Severini, Silvia and Sabet, Masoud Jalili and Kassner, Nora and Ma, Chunlan and Schmid, Helmut and Martins, Andr{\'e} FT and Yvon, Fran{\c{c}}ois and others},
  booktitle={Proceedings of the 61st Annual Meeting of the Association for Computational Linguistics (Volume 1: Long Papers)},
  pages={1082--1117},
  year={2023}
}

@inproceedings{etxaniz2024multilingual,
  title={Do multilingual language models think better in English?},
  author={Etxaniz, Julen and Azkune, Gorka and Soroa, Aitor and de Lacalle, Oier Lopez and Artetxe, Mikel},
  booktitle={Proceedings of the 2024 Conference of the North American Chapter of the Association for Computational Linguistics: Human Language Technologies (Volume 2: Short Papers)},
  pages={550--564},
  year={2024}
}

@inproceedings{zhang2025shifcon,
  title={ShifCon: Enhancing Non-Dominant Language Capabilities with a Shift-based Multilingual Contrastive Framework},
  author={Zhang, Hengyuan and Shang, Chenming and Wang, Sizhe and Zhang, Dongdong and Yu, Yiyao and Yao, Feng and Sun, Renliang and Yang, Yujiu and Wei, Furu},
  booktitle={Proceedings of the 63rd Annual Meeting of the Association for Computational Linguistics (Volume 1: Long Papers)},
  pages={4818--4841},
  year={2025}
}

@inproceedings{rei2022cometkiwi,
  title={CometKiwi: IST-unbabel 2022 submission for the quality estimation shared task},
  author={Rei, Ricardo and Treviso, Marcos and Guerreiro, Nuno M and Zerva, Chrysoula and Farinha, Ana C and Maroti, Christine and De Souza, Jos{\'e} GC and Glushkova, Taisiya and Alves, Duarte and Coheur, Luisa and others},
  booktitle={Proceedings of the Seventh Conference on Machine Translation (WMT)},
  pages={634--645},
  year={2022}
}

@inproceedings{holtermann2024evaluating,
  title={Evaluating the elementary multilingual capabilities of large language models with multiq},
  author={Holtermann, Carolin and R{\"o}ttger, Paul and Dill, Timm and Lauscher, Anne},
  booktitle={Findings of the Association for Computational Linguistics: ACL 2024},
  pages={4476--4494},
  year={2024}
}

@inproceedings{hangya2022improving,
  title={Improving low-resource languages in pre-trained multilingual language models},
  author={Hangya, Viktor and Saadi, Hossain Shaikh and Fraser, Alexander},
  booktitle={Proceedings of the 2022 Conference on Empirical Methods in Natural Language Processing},
  pages={11993--12006},
  year={2022}
}

@inproceedings{singh2023h,
  title={H-AES: Towards automated essay scoring for Hindi},
  author={Singh, Shubhankar and Pupneja, Anirudh and Mital, Shivaansh and Shah, Cheril and Bawkar, Manish and Gupta, Lakshman Prasad and Kumar, Ajit and Kumar, Yaman and Gupta, Rushali and Shah, Rajiv Ratn},
  booktitle={Proceedings of the AAAI Conference on Artificial Intelligence},
  volume={37},
  number={13},
  pages={15955--15963},
  year={2023}
}

@inproceedings{dubossarsky2024strengthening,
  title={Strengthening the wic: New polysemy dataset in hindi and lack of cross lingual transfer},
  author={Dubossarsky, Haim and Dairkee, Farheen},
  booktitle={Proceedings of the 2024 Joint International Conference on Computational Linguistics, Language Resources and Evaluation (LREC-COLING 2024)},
  pages={15341--15349},
  year={2024}
}

@inproceedings{sharma2025hi,
  title={Hi-GEC: Hindi grammar error correction in low resource scenario},
  author={Sharma, Ujjwal and Bhattacharyya, Pushpak},
  booktitle={Proceedings of the 31st International Conference on Computational Linguistics},
  pages={6063--6075},
  year={2025}
}

@inproceedings{anoop2021unsupervised,
  title={Unsupervised domain adaptation schemes for building ASR in low-resource languages},
  author={Anoop, Chandran Savithri and Prathosh, AP and AG, Ramakrishnan},
  booktitle={2021 IEEE Automatic Speech Recognition and Understanding Workshop (ASRU)},
  pages={342--349},
  year={2021},
  organization={IEEE}
}

@article{zhu2024multilingual,
  title={Multilingual large language models: A systematic survey},
  author={Zhu, Shaolin and Xu, Shaoyang and Sun, Haoran and Pan, Leiyu and Cui, Menglong and Du, Jiangcun and Jin, Renren and Branco, Ant{\'o}nio and Xiong, Deyi and others},
  journal={arXiv preprint arXiv:2411.11072},
  year={2024}
}

@inproceedings{zhu2024landermt,
  title={Landermt: Dectecting and routing language-aware neurons for selectively finetuning llms to machine translation},
  author={Zhu, Shaolin and Pan, Leiyu and Li, Bo and Xiong, Deyi},
  booktitle={Proceedings of the 62nd Annual Meeting of the Association for Computational Linguistics (Volume 1: Long Papers)},
  pages={12135--12148},
  year={2024}
}

@article{zhu2025overcoming,
  title={Overcoming language barriers via machine translation with sparse mixture-of-experts fusion of large language models},
  author={Zhu, Shaolin and Pan, Leiyu and Jian, Dong and Xiong, Deyi},
  journal={Information Processing \& Management},
  volume={62},
  number={3},
  pages={104078},
  year={2025},
  publisher={Elsevier}
}

@inproceedings{li202510m,
  title={MIT-10M: A large scale parallel corpus of multilingual image translation},
  author={Li, Bo and Zhu, Shaolin and Wen, Lijie},
  booktitle={Proceedings of the 31st International Conference on Computational Linguistics},
  pages={5154--5167},
  year={2025}
}

@article{zhang2024lingualift,
  title={Lingualift: an effective two-stage instruction tuning framework for low-resource language tasks},
  author={Zhang, Hongbin and Chen, Kehai and Bai, Xuefeng and Xiang, Yang and Zhang, Min},
  journal={arXiv e-prints},
  pages={arXiv--2412},
  year={2024}
}

@article{zhang2026mitigating,
  title={Mitigating Translationese Bias in Multilingual LLM-as-a-Judge via Disentangled Information Bottleneck},
  author={Zhang, Hongbin and Chen, Kehai and Bai, Xuefen and Pan, Youcheng and Xiang, Yang and Wang, Jinpeng and Zhang, Min},
  journal={arXiv preprint arXiv:2603.10351},
  year={2026}
}

@inproceedings{chen2025benchmarking,
  title={Benchmarking llms for translating classical chinese poetry: Evaluating adequacy, fluency, and elegance},
  author={Chen, Andong and Lou, Lianzhang and Chen, Kehai and Bai, Xuefeng and Xiang, Yang and Yang, Muyun and Zhao, Tiejun and Zhang, Min},
  booktitle={Proceedings of the 2025 Conference on Empirical Methods in Natural Language Processing},
  pages={33007--33024},
  year={2025}
}

@inproceedings{zeng2023scomoe,
  title={SCoMoE: Efficient Mixtures of Experts with Structured Communication.},
  author={Zeng, Zhiyuan and Xiong, Deyi},
  booktitle={ICLR},
  year={2023}
}

@article{pan2025advancing,
  title={Advancing Large Language Models for Tibetan with Curated Data and Continual Pre-Training},
  author={Pan, Leiyu and Xiong, Bojian and Yang, Lei and Jin, Renren and Zhang, Shaowei and Chen, Yue and Shi, Ling and Zhou, Jiang and Wu, Junru and Wang, Zhen and others},
  journal={arXiv preprint arXiv:2507.09205},
  year={2025}
}

\clearpage
\appendix
\section{Appendix}
\begin{figure*}
    \centering
    \includegraphics[width=\textwidth]{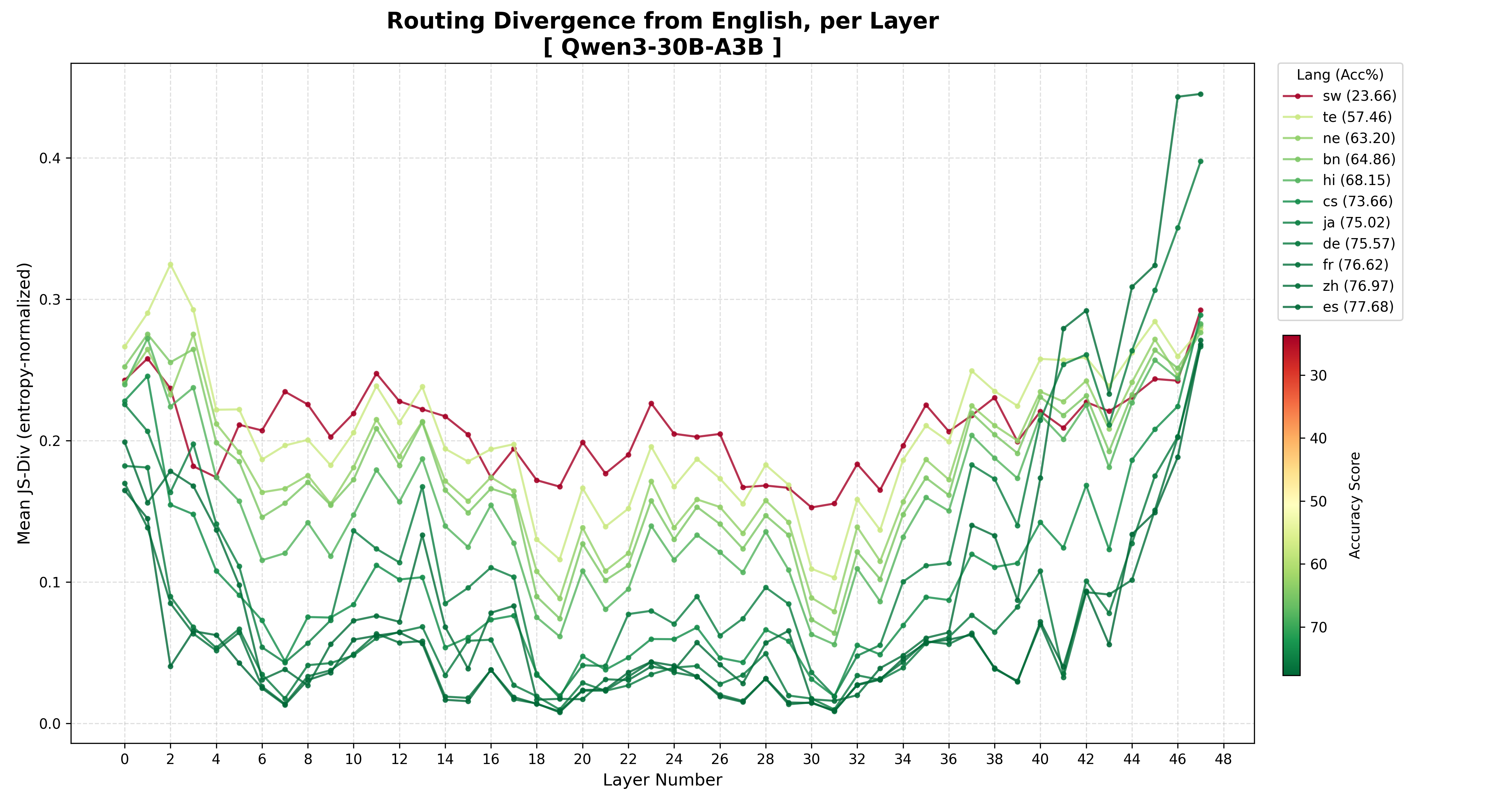}
    \caption{Visualization of routing divergence for selected languages relative to English across model layers based on Qwen3‑30B‑A3B on the Global-MMLU dataset.}
    \label{fig:routing_divergence_qwen}
\end{figure*}

\begin{figure*}
    \centering
    \includegraphics[width=\textwidth]{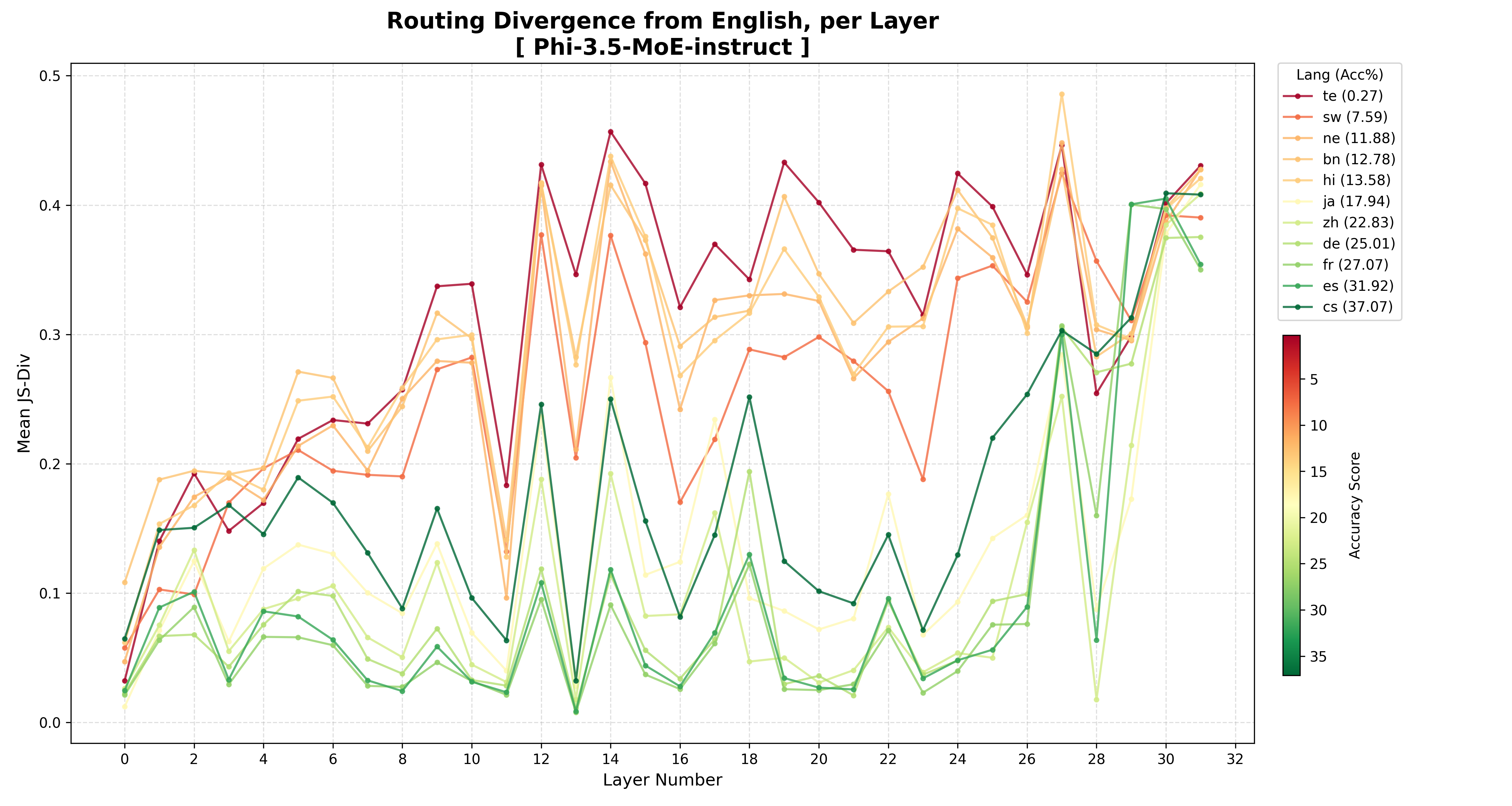}
    \caption{Visualization of routing divergence for selected languages relative to English across model layers based on Phi-3.5-MoE-instruct on the Global-MMLU dataset.}
    \label{fig:routing_divergence_phi}
\end{figure*}

\subsection{Details of Languages}
\label{app:detail_languages}
In our experiments, we designate en and zh as high-resource anchor languages. We select hi, ne, bn, te and sw as the representative low-resource target languages. The full list of languages and their corresponding ISO 639 codes is provided in Table \ref{tab:detail_languages}.

\begin{table}[h!]
\centering
\begin{tabular}{cc}
\toprule
\textbf{ISO 639} & \textbf{Language} \\
\midrule
hi & Hindi \\
ne & Nepali \\
bn & Bengali \\
te & Telugu \\
sw& Swahili \\
en & English \\
zh & Chinese \\
\bottomrule
\end{tabular}
\caption{ISO 639 language codes and names.}
\label{tab:detail_languages}
\end{table}

\subsection{Detailed Analysis of Routing Divergence and Layer Selection}
\label{app:layer_analysis}

Figures \ref{fig:routing_divergence_qwen} and \ref{fig:routing_divergence_phi} visualize the layer-wise routing divergence (measured by JS divergence) between English and various target languages for Qwen3-30B-A3B and Phi-3.5-MoE-instruct. A core finding is that both models exhibit a characteristic U-shaped distribution relative to the model depth: divergence is significantly higher in the initial layers (which handle shallow linguistic features) and the final layers (which specialize in language-specific token generation), while the intermediate layers form a stable, lower-divergence valley. This U-shaped trend suggests that the mid-range experts function as a language-agnostic semantic reasoning engine, making them the ideal targets for semantically anchored alignment.

Based on these empirical observations, we selectively apply the routing alignment loss $\mathcal{L}_\text{JS}$ to the layers corresponding to these U-shaped valleys. For Qwen3-30B-A3B, the divergence reaches its nadir between layer 7 and 34, which we define as our alignment range. Phi-3.5-MoE-instruct similarly follows this U-shaped trend; despite its higher local volatility, the divergence is consistently lower within the intermediate routing logic of layers 13 to 26. By focusing SARA on these specific ranges, we ensure the supervision signal from high-resource anchors rectifies the core reasoning logic while avoiding the noise inherent in the model's shallow input processing and deep output generation stages.

\subsection{Sensitivity Analysis of $\lambda_\text{JS}$}
\label{app:lambda_js}

To evaluate the impact of the routing alignment strength, we conduct a sensitivity analysis on the hyperparameter $\lambda_\text{JS}$. This parameter controls the weight of the Jensen–Shannon divergence loss within the total training objective ($\mathcal{L}_{total}$). We vary $\lambda_\text{JS}$ across $\{0.5, 1.0, 1.5, 2.0\}$ and evaluate the model's performance on the Global-MMLU benchmark to observe how the magnitude of the alignment constraint affects multilingual reasoning.

\begin{figure}
    \centering
    \includegraphics[width=\linewidth]{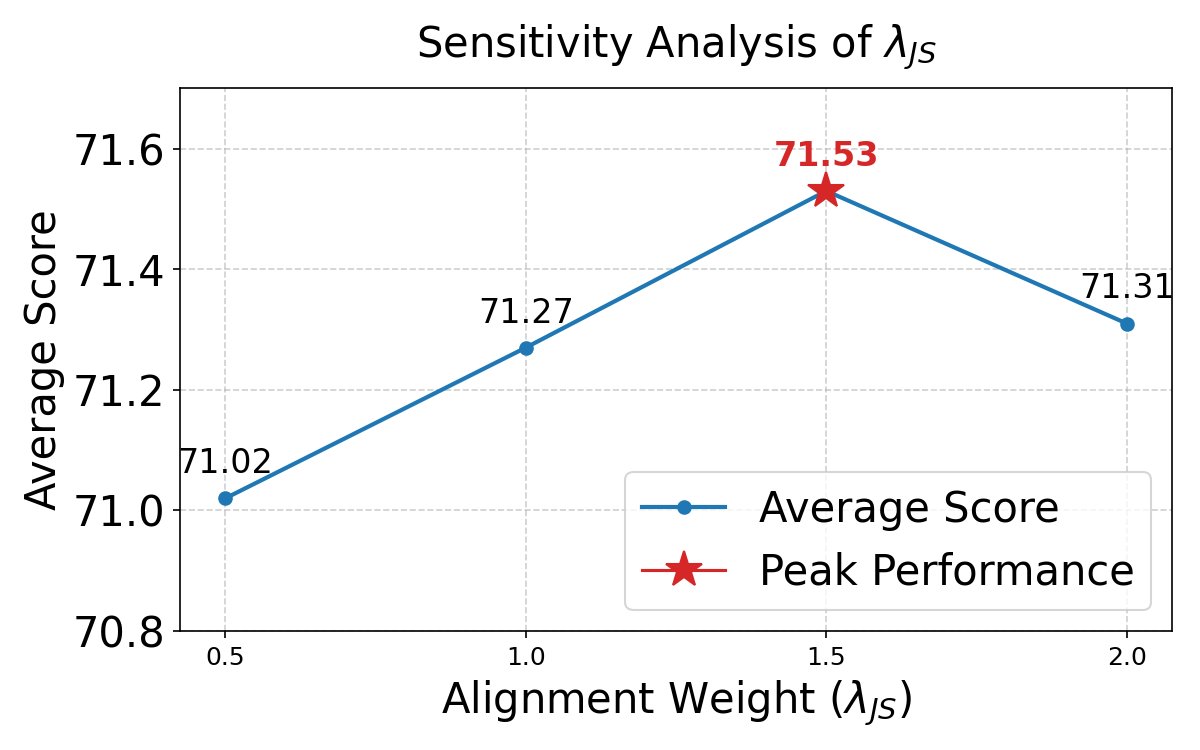}
    \caption{Average score as $\lambda_\text{JS}$ changes.}
    \label{fig:lambda_sensitivity}
\end{figure}

The results are summarized in Table \ref{tab:sensitivity_lambda} and Figure \ref{fig:lambda_sensitivity}. We observe that the model performance generally benefits from a moderate alignment strength. When $\lambda_\text{JS}$ is low (0.5 or 1.0), the alignment signal is insufficient to fully transfer semantic patterns, resulting in lower average scores. Conversely, an overly strict constraint ($\lambda_\text{JS}=2.0$) slightly hampers the model's flexibility, leading to a performance drop. The setting $\lambda_\text{JS} = 1.5$ achieves the highest average Global-MMLU accuracy of 71.53, offering the optimal balance between semantic consistency and task-specific generation across the diverse set of languages.

\begin{table*}[t]
\centering
\footnotesize
\renewcommand{\arraystretch}{0.88}
\setlength{\tabcolsep}{12pt}
\begin{tabular}{lcccc}
\toprule
\textbf{Language} & \textbf{$\lambda_\text{JS}=2.0$} & \textbf{$\lambda_\text{JS}=1.5$} & \textbf{$\lambda_\text{JS}=1.0$} & \textbf{$\lambda_\text{JS}=0.5$} \\
\midrule
hi    & 74.33 & \textbf{74.66} & 74.19 & 73.81 \\
ne   & \textbf{72.60} & \textbf{72.60} & 72.40 & 72.52 \\
bn  & 72.33 & \textbf{72.70} & 72.48 & 72.10 \\
te   & 66.40 & 65.76 & \textbf{66.46} & 65.79 \\
sw  & 59.76 & \textbf{61.00} & 59.71 & 59.41 \\
en  & 82.46 & 82.45 & 82.39 & \textbf{82.51} \\
\midrule
\textit{Avg.} & 71.31 & \textbf{71.53} & 71.27 & 71.02 \\
\bottomrule
\end{tabular}
\caption{Sensitivity analysis of the routing alignment weight $\lambda_\text{JS}$ on the Global-MMLU benchmark. We report the performance on individual languages and the overall average. The results indicate that $\lambda_\text{JS}=1.5$ yields the best balance.}
\label{tab:sensitivity_lambda}
\end{table*}

\subsection{Statistical Significance Testing}
\label{app: Statistical Significance Testing}
To rigorously evaluate the performance improvements of SARA relative to the standard full fine-tuning (FFT) baseline, we conducted one-tailed paired t-tests using performance scores across all evaluated languages as paired samples.

\subsubsection{Hypothesis Formulation}
Given that SARA is explicitly designed to enhance cross-lingual generalization compared to standard fine-tuning, we employed a directional hypothesis:
$$H_0: \mu_D \le 0$$
$$H_1: \mu_D > 0$$
where $D_i = X_{\text{SARA},i} - X_{\text{FFT},i}$ represents the performance difference for a specific language $i$. The t-statistic is calculated as:
$$t = \frac{\bar{D}}{SE_{\bar{D}}}, \quad df = n-1$$
The one-tailed $p$-value is derived from the upper tail of the t-distribution with $n-1$ degrees of freedom, representing the probability of observing a positive mean difference at least as large as $\bar{D}$ under the null hypothesis.

\subsubsection{Results and Analysis}
The statistical results across three multilingual benchmarks—Global-MMLU, BELEBELE, and MGSM—for the Qwen3-30B-A3B and Phi-3.5-MoE-instruct models are summarized in the following table:

\begin{table}[h]
\centering
\resizebox{\columnwidth}{!}{

\begin{tabular}{llcc}
\toprule
\textbf{Benchmark} & \textbf{Model} & \textbf{t-statistic} & \textbf{p-value} \\
\midrule
Global-MMLU & Qwen3-30B-A3B & 4.0784 & \textbf{0.0096} \\
& Phi-3.5-MoE-instruct & 2.4705 & \textbf{0.0565} \\
\midrule
BELEBELE & Qwen3-30B-A3B & 0.4200 & 0.6904 \\
& Phi-3.5-MoE-instruct & 2.0100 & 0.1008 \\
\midrule
MGSM & Qwen3-30B-A3B & 0.1525 & 0.8885 \\
& Phi-3.5-MoE-instruct & 1.4606 & 0.2403 \\
\bottomrule
\end{tabular}
}
\caption{One-tailed paired t-test results comparing SARA and FFT.}
\end{table}

\begin{itemize}[leftmargin=*]
    \item \textbf{Global-MMLU}: This benchmark provides the strongest statistical evidence due to its large evaluation set of approximately 14,000 samples per language. SARA demonstrates statistically supported improvements, particularly for the Qwen3 model($p < 0.01$).
    \item \textbf{BELEBELE and MGSM}: These benchmarks consist of significantly fewer test samples—approximately 900 and 250 per language, respectively. While SARA maintains positive performance trends, the lack of statistical significance in these instances is primarily due to reduced statistical power from smaller sample sizes rather than inconsistent model behavior.
\end{itemize}

\subsection{Results on Other LLMs}
\label{app:phimoe}
We also fine-tuned the Phi-3.5-MoE-instruct model using our proposed method and compared it against baseline fine-tuning approaches. The results, presented in Table \ref{tab:phi35_results}, demonstrate that our method achieves state-of-the-art performance across multiple datasets and languages. This finding underscores the generalizability of our approach across different model architectures. Notably, the performance gain observed on Phi-3.5-MoE-instruct is more pronounced than that on Qwen3-30B-A3B, suggesting that our method may yield greater improvements on models with relatively weaker multilingual capabilities.

\begin{table*}[t]
\centering
\footnotesize
\renewcommand{\arraystretch}{0.88}
\setlength{\tabcolsep}{8pt}

\begin{tabular}{ll ccccccc}
\toprule
\textbf{Benchmark} & \textbf{Method} & \textbf{hi} & \textbf{ne} & \textbf{bn} & \textbf{te} & \textbf{sw} & \textbf{en} & \textbf{Avg.} \\
\midrule
\multirow{5}{*}{\textbf{Global-MMLU}} & Vanilla LM & 13.58 & 11.88 & 12.78 & 0.27 & 7.59 & 49.16 & 15.88 \\
& FFT & 42.14 & 37.39 & 27.99 & 15.96 & 45.42 & 80.86 & 41.63 \\
& AES & 40.55 & 36.12 & 26.54 & 12.45 & 43.20 & 80.15 & 39.84 \\
& ShifCon & 43.15 & 37.82 & 28.35 & 17.55 & 45.54 & 80.92 & 42.22 \\
& \textbf{SARA (Ours)} & \textbf{44.14} & \textbf{38.25} & \textbf{28.70} & \textbf{19.09} & \textbf{45.66} & \textbf{80.98} & \textbf{42.80} \\
\midrule
\multirow{5}{*}{\textbf{BELEBELE}} & Vanilla LM & 10.33 & 1.11 & 24.44 & 0.00 & 0.56 & 51.00 & 14.57 \\
& FFT & 68.89 & 64.56 & 55.56 & 0.00 & 58.22 & \textbf{93.78} & 56.84 \\
& AES & 67.25 & 63.15 & 53.25 & 0.00 & 56.15 & 92.55 & 55.39 \\
& ShifCon & 70.15 & 64.85 & 58.45 & 0.05 & 59.85 & 93.65 & 57.83 \\
& \textbf{SARA (Ours)} & \textbf{71.22} & \textbf{65.00} & \textbf{61.56} & \textbf{0.11} & \textbf{61.67} & 93.44 & \textbf{58.83} \\
\midrule
\multirow{5}{*}{\textbf{MGSM}} & Vanilla LM & - & - & 9.20 & 0.00 & 11.20 & 65.20 & 21.40 \\
& FFT & - & - & 54.80 & 44.00 & \textbf{55.60} & 66.40 & 55.20 \\
& AES & - & - & 53.20 & 42.40 & 52.00 & 65.60 & 53.30 \\
& ShifCon & - & - & 56.40 & 45.20 & 53.60 & 68.00 & 55.80 \\
& \textbf{SARA (Ours)} & - & - & \textbf{58.00} & \textbf{46.00} & 54.00 & \textbf{69.20} & \textbf{56.80} \\
\bottomrule
\end{tabular}
\caption{Performance comparison of Phi-3.5-MoE-instruct on Global-MMLU, BELEBELE, and MGSM using \textbf{English} as the semantic anchor.}
\label{tab:phi35_results}
\end{table*}

\subsection{Ablation Variant Definitions}
\label{sec:ablation_variants}

To further analyze the design choices in SARA, we define several controlled ablation variants by systematically varying the source of routing priors, anchor language selection, and layer alignment strategies.

\begin{itemize}[leftmargin=*]
    \item \textbf{-q-en-s (Ours):} Uses internal inference traces generated by Qwen3-30B-A3B (\textit{-q}) and English anchors (\textit{en}) on the selected intermediate layers (\textit{-s}).
    \item \textbf{-g-en-s:} Uses external data generated and translated by GPT-5 mini (\textit{-g}) to obtain routing priors through forward propagation on Qwen3-30B-A3B, serving as the English anchors (\textit{en}) for the selected intermediate layers (\textit{-s}) to evaluate the necessity of self-anchoring.
    \item \textbf{-q-en-r:} Uses internal inference traces generated by Qwen3-30B-A3B (\textit{-q}) and English anchors (\textit{en}) but aligns a random subset of layers (\textit{-r}) instead of the targeted intermediate ones, keeping the total number of aligned layers consistent with the \textit{-s} setting.
    \item \textbf{-q-en-a:} Uses internal inference traces generated by Qwen3-30B-A3B (\textit{-q}) and English anchors (\textit{en}) while applying alignment across all layers (\textit{-a}) of the model.
    \item \textbf{-q-sw-s:} Uses internal inference traces generated by Qwen3-30B-A3B (\textit{-q}) on the selected intermediate layers (\textit{-s}) but utilizes low-resource Swahili (\textit{sw}) as the semantic anchor to test the pivot language quality.
\end{itemize}

\subsection{Prompt Templates}
\label{app:prompt_templates}
To evaluate the multilingual capabilities of LLMs across multiple benchmarks (Global-MMLU, BELEBELE, and MGSM), we developed standardized prompt templates in English. To ensure cross-lingual consistency and minimize instruction bias, we utilized GPT-5 mini to translate these core instructions into our target languages, maintaining the semantic structure while adapting to the natural linguistic flow of each language.

To facilitate objective and scalable evaluation, we mandated that models enclose their final answers within a \texttt{\textbackslash boxed\{\}} command. We subsequently employed regular expressions to programmatically parse the model outputs and extract the final answers for performance scoring. The base English templates for each benchmark are detailed in Table \ref{tab:prompt-template-en-global-mmlu}, \ref{tab:prompt-template-en-belebele}, and \ref{tab:prompt-template-en-mgsm}.

\begin{table}[h]
\centering
\begin{tcolorbox}[
    colback=gray!5,
    colframe=black,
    arc=0mm,
    boxrule=0.5pt,
    fontupper=\small,
    left=10pt, right=10pt, top=10pt, bottom=10pt
]
Below is a single-choice question with four options (A, B, C, D). Only one option is correct. \\
\\
\textbf{Question:} \\
\{\texttt{question}\} \\
\\
\textbf{Options:} \\
A) \{\texttt{A}\} \\
B) \{\texttt{B}\} \\
C) \{\texttt{C}\} \\
D) \{\texttt{D}\} \\
\\
Analyze the question carefully and answer it. At the end of your response, clearly indicate your final answer by enclosing the correct choice in \textbackslash\texttt{boxed\{\}}.
\end{tcolorbox}
\caption{Prompt template used for Global-MMLU.}
\label{tab:prompt-template-en-global-mmlu}
\end{table}

\begin{table}[h]
\centering
\begin{tcolorbox}[colback=gray!5, colframe=black, arc=0mm, boxrule=0.5pt, fontupper=\small]
Below is a passage followed by a single-choice question with four options (A, B, C, D). Only one option is correct based on the passage. \\
\\
\textbf{Passage:} \\
\{\texttt{flores\_passage}\} \\
\\
\textbf{Question:} \\
\{\texttt{question}\} \\
\\
\textbf{Options:} \\
A) \{\texttt{A}\} \\
B) \{\texttt{B}\} \\
C) \{\texttt{C}\} \\
D) \{\texttt{D}\} \\
\\
Analyze the passage and question carefully and answer. At the end of your response, clearly indicate your final answer by enclosing the correct choice in \textbackslash\texttt{boxed\{\}}.
\end{tcolorbox}
\caption{Prompt template used for BELEBELE.}
\label{tab:prompt-template-en-belebele}
\end{table}

\begin{table}[h]
\centering
\begin{tcolorbox}[
    colback=gray!5,
    colframe=black,
    arc=0mm,
    boxrule=0.5pt,
    fontupper=\small,
    left=10pt, right=10pt, top=10pt, bottom=10pt
]
Below is a question. \\
\\
\textbf{Question:} \\
\{\texttt{question}\} \\
\\
Please analyze the question and solve it. At the end of your response, clearly indicate your final answer by enclosing it in \textbackslash\texttt{boxed\{\}}.
\end{tcolorbox}
\caption{Prompt template for MGSM.}
\label{tab:prompt-template-en-mgsm}
\end{table}

\end{document}